\pdfoutput=1

\documentclass[11pt]{article}

\usepackage[]{ACL2023}

\usepackage{times}
\usepackage{latexsym}

\usepackage[T1]{fontenc}

\usepackage[utf8]{inputenc}

\usepackage{microtype}

\usepackage{inconsolata}

\usepackage{url} %
\usepackage{booktabs} %
\usepackage{amsfonts} %
\usepackage{nicefrac} %
\usepackage{amsmath}
\usepackage{graphicx}
\usepackage{bm, upgreek}
\usepackage{amssymb}
\usepackage{array, booktabs}
\usepackage{multirow}
\usepackage{mathtools}
\usepackage{arydshln}
\usepackage{tabularx}
\usepackage{color}
\usepackage{soul}
\usepackage{lipsum}
\usepackage{amsthm}
\usepackage{caption}
\usepackage{xcolor}
\usepackage{pifont}%
\usepackage{comment}
\usepackage{cleveref}
\usepackage{enumitem}
\usepackage{subcaption}
\usepackage[most]{tcolorbox}
\usepackage[ruled,vlined]{algorithm2e}
\crefformat{section}{\S#2#1#3} %
\crefformat{subsection}{\S#2#1#3}
\crefformat{subsubsection}{\S#2#1#3}
\usepackage{tablefootnote}

\definecolor{green}{RGB}{27,158,119}
\definecolor{orange}{RGB}{217,95,2}
\definecolor{mycolor3}{RGB}{117,112,179}
\definecolor{mycolor5}{RGB}{228,26,28}
\definecolor{mycolor4}{RGB}{55,126,184}
\definecolor{mycolor6}{RGB}{77,175,74}
\definecolor{mycolor7}{RGB}{152,78,163}

\newcommand{\ours}{\textsc{TouR}}

\newcommand{\topkret}[1]{\mathcal{C}_{1:k}^{#1}}
\newcommand{\toppret}{\mathcal{C}_{\text{hard}}^{q}}
\newcommand{\pkctq}{P_k(\tilde{c}|q_t)}
\newcommand{\pctq}{P(\tilde{c}|q_t)}
\newcommand{\pkcq}{P_k(c|q_t)}
\newcommand{\cross}[1]{\phi#1}

\newcommand\ti[1]{\textit{#1}}

\newcommand{\cb}{\mathbf{c}}

\newcommand{\qb}{\mathbf{q}}

\newcommand{\wb}{\mathbf{w}}

\newtcbox{\hlprimarytab}{on line, rounded corners, box align=base, colback=red!10,colframe=white,size=fbox,arc=3pt, before upper=\strut, top=-2pt, bottom=-4pt, left=-2pt, right=-2pt, boxrule=0pt}
\newtcbox{\hlsecondarytab}{on line, box align=base, colback=blue!10,colframe=white,size=fbox,arc=3pt, before upper=\strut, top=-2pt, bottom=-4pt, left=-2pt, right=-2pt, boxrule=0pt}
\newtcbox{\hlthirdtab}{on line, box align=base, colback=gray!10,colframe=white,size=fbox,arc=3pt, before upper=\strut, top=-2pt, bottom=-4pt, left=-2pt, right=-2pt, boxrule=0pt}
\newtcbox{\hlfourthtab}{on line, box align=base, colback=blue!20,colframe=white,size=fbox,arc=3pt, before upper=\strut, top=-2pt, bottom=-4pt, left=-2pt, right=-2pt, boxrule=0pt}

\newcommand{\dashifted}{\raisebox{0.5\depth}{\tiny$\downarrow$}}
\newcommand{\muashifted}{\raisebox{0.5\depth}{\tiny$\uparrow$}}
\newcommand{\uashifted}{\raisebox{0.5\depth}{\tiny$\uparrow$}}
\newcommand{\nashifted}{\raisebox{0.5\depth}{\tiny$$}}
\newcommand{\da}[1]{{\small\hlprimarytab{\dashifted{#1}}}}
\newcommand{\ua}[1]{{\small\hlsecondarytab{\uashifted{#1}}}}
\newcommand{\mua}[1]{{\small\hlfourthtab{\muashifted{#1}}}}
\newcommand{\na}[1]{{\small\hlthirdtab{\nashifted{#1}}}}

\definecolor{readablered}{RGB}{170,0,0}

\title{Optimizing Test-Time Query Representations for Dense Retrieval}

\author{
    Mujeen Sung$^{1}$\quad Jungsoo Park$^{1}$\quad Jaewoo Kang$^{1}$\quad Danqi Chen$^{2}$\quad Jinhyuk Lee$^{3}$\Thanks{~Work partly done while visiting Princeton University.} \\
    Korea University$^{1}$\quad Princeton University$^{2}$\quad Google Research$^{3}$\\
    \texttt{\{mujeensung,jungsoo\_park,kangj\}@korea.ac.kr} \\
    \texttt{danqic@cs.princeton.edu \quad jinhyuklee@google.com} 
}

\begin{document}
\maketitle
\begin{abstract}
Recent developments of dense retrieval rely on quality representations of queries and contexts from pre-trained query and context encoders.
In this paper, we introduce \ours{} (\textbf{T}est-Time \textbf{O}ptimization of Q\textbf{u}ery \textbf{R}epresentations), which further optimizes \textit{instance-level} query representations guided by signals from test-time retrieval results.
We leverage a cross-encoder re-ranker to provide fine-grained \textit{pseudo labels} over retrieval results and iteratively optimize query representations with gradient descent.
Our theoretical analysis reveals that \ours{} can be viewed as a generalization of the classical Rocchio algorithm for pseudo relevance feedback, and we present two variants that leverage pseudo-labels as hard binary or soft continuous labels.
We first apply \ours{} on phrase retrieval with our proposed phrase re-ranker, and also evaluate its effectiveness on passage retrieval with an off-the-shelf re-ranker.
\ours{} greatly improves end-to-end open-domain question answering accuracy, as well as passage retrieval performance.
\ours{} also consistently improves direct re-ranking  by up to 2.0\% while running {1.3--2.4$\times$} faster with an efficient implementation.\footnote{Our code is available at \url{https://github.com/dmis-lab/TouR}.}

\end{abstract}

\section{Introduction}

\begin{figure}[t]
\centering
\includegraphics[width=0.95\columnwidth]{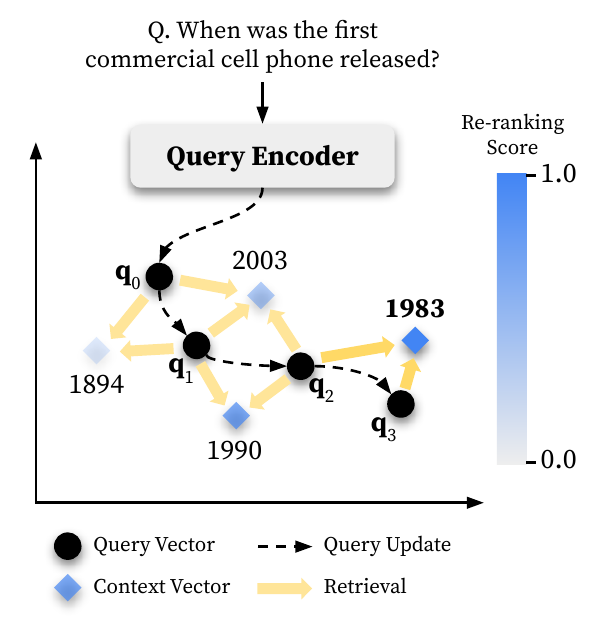}
\caption{
An overview of test-time optimization of query representations (\ours).
Given the initial representation of a test query $\mathbf{q}_0$, \ours{} iteratively optimizes its representation (e.g., $\mathbf{q}_0 \rightarrow \mathbf{q}_1 \rightarrow \mathbf{q}_2 \rightarrow \mathbf{q}_3$) based on top-$k$ retrieval results.
The figure shows how each query vector retrieves new context vectors and updates its representation to find the gold answer (e.g., 1983).
Our cross-encoder re-ranker provides a relevance score for each top retrieval result making the query representation closer to the final answer.}
\vspace{-0.2cm}
\label{fig:overview}
\end{figure}

Recent progress in pre-trained language models gave birth to dense retrieval, which typically learns dense representations of queries and contexts in a contrastive learning framework.
By overcoming the term mismatch problem, dense retrieval has been shown to be more effective than sparse retrieval in open-domain question answering (QA)~\citep{lee2019latent,karpukhin2020dense,lee2021learning} and information retrieval~\citep{khattab2020colbert, xiong2020approximate}.

Dense retrieval often uses a dual encoder architecture, which enables the pre-computation of context representations while the query representations are directly computed from the trained encoder during inference.
However, directly using trained query encoders often fails to retrieve the relevant context~\citep{thakur2021beir,sciavolino2021simple} as many test queries are unseen during training.

In this paper, we introduce \ours, which further optimizes instance-level query representations at test time for dense retrieval.
Specifically, we treat each test query as a single data point and iteratively optimize its representation.
This resembles the query-side fine-tuning proposed for phrase retrieval~\citep{lee2021learning}, which fine-tunes the query encoder over \ti{training} queries in a new domain.
Instead, we fine-tune query representations for each \ti{test} query.
Cross-encoders are known to exhibit better generalization ability in unseen distributions compared to dual encoders~\citep{rosa2022defense}.
{Accordingly,} we leverage cross-encoder re-rankers~\citep{nogueira2019passage, fajcik2021r2} to provide \textit{pseudo relevance labels} on intermediate retrieval results and then iteratively optimize query representations using gradient descent.
For phrase retrieval, we also develop a cross-encoder phrase re-ranker, which has not been explored in previous studies.

We theoretically show that our framework can be viewed as a generalized version of the Rocchio algorithm for pseudo relevance feedback~(PRF;~\citealp{rocchio1971relevance}), which is commonly used in information retrieval to improve query representations.
While most PRF techniques assume that the top-ranked results are equally pseudo-relevant, our method dynamically labels the top results and updates the query representations accordingly.
We leverage our pseudo labels as either hard binary or soft continuous labels in two instantiations of our method, respectively.
Lastly, to reduce computational overhead, we present an efficient implementation of \ours{}, which significantly improves its runtime efficiency.

We apply \ours{} on phrase~\citep{lee2021learning} and passage retrieval~\citep{karpukhin2020dense} for open-domain QA.
{Experiments show that \ours{} consistently improves performance in both tasks, even when the query distribution changes greatly.
Specifically, \ours{} improves the end-to-end open-domain QA accuracy by up to 10.7\%, while also improving the accuracy of the top-20 passage retrieval by up to 8.3\% compared to baseline retrievers.}
{\ours{} requires only a handful of top-$k$ candidates to perform well, which enables \ours{} to run up to {1.3--2.4$\times$} faster than the direct application of re-ranker with our efficient implementation while consistently improving the performance by up to 2.0\%.}
The ablation study further shows the effectiveness of each component, highlighting the importance of fine-grained relevance signals.

\section{Background}

\subsection{Dense Retrieval}
Dense retrieval typically uses query and context encoders---$E_\text{q}(\cdot)$ and $E_\text{c}(\cdot)$---for representing queries and contexts, respectively~\citep{lee2019latent,karpukhin2020dense}.
In this work, we focus on improving phrase or passage retrievers for open-domain QA.
The similarity of a query $q$ and a context $c$ is computed based on the inner product between their dense representations:
\begin{equation}
    \text{sim}(q, c) = E_q(q)^{\intercal}E_c(c) = \mathbf{q}^{\intercal}\mathbf{c}.
\end{equation}

Dense retrievers often use the contrastive learning framework to train encoders $E_q$ and $E_c$.
After training the encoders, top-$k$ results are retrieved from a set of contexts $\mathcal{C}$:
\begin{equation}
    \topkret{q} = [c_1, \dots, c_k] = \text{top-}k_{c \in \mathcal{C}} \text{sim}(q, c),
\end{equation}
where the top-$k$ operator returns a sorted list of contexts by their similarity score $\text{sim}(q, c)$ in descending order, i.e., $\text{sim}(q, c_1) \geq \dots \geq \text{sim}(q, c_k)$.
Dense retrievers aim to maximize the probability that a relevant context $c^*$ exists (or is highly ranked) in the top results.

\subsection{Query-side Fine-tuning}
After training the query and context encoders, the context representations $\{\cb \mid c \in \mathcal{C}\}$ are typically pre-computed for efficient retrieval while the query representations $\qb$ are directly computed from the query encoder during inference.
However, using the dense representations of queries as is often fails to retrieve relevant contexts, especially when the test query distribution is different from the one seen during training.

To mitigate the problem, \citet{lee2021learning} propose to fine-tune the query encoder on the retrieval results of training queries $\{q \mid q \in \mathcal{Q}_\text{train}\}$ over the entire corpus $\mathcal{C}$.
For phrase retrieval (i.e., $c$ denotes a phrase), they maximize the marginal likelihood of relevant phrases in the top-$k$ results:
\begin{equation}\label{eq:qsft}
\begin{split}
    &\mathcal{L}_\text{query} = -\sum_{q \in \mathcal{Q}_\text{train}}\log \smashoperator{\sum_{c \in \topkret{q}, c=c^*}} P_k(c|q),
\end{split}
\end{equation}
where $P_k(c|q) = \frac{\exp({\text{sim}(q, c)})}{\sum_{i=1}^k \exp({\text{sim}(q, c_i)})}$ and $c=c^*$ checks whether each context matches the gold context $c^*$ or not.
Note that $c^*$ is always given for training queries.
The query-side fine-tuning significantly improves performance and provides a means of efficient transfer learning when there is a query distribution shift.
In this work, compared to training on entire training queries as in Eq.~\eqref{eq:qsft}, we treat each test query $q \in \mathcal{Q}_\text{test}$ as a single data point to train on and optimize instance-level query representations at test time.
This is in contrast to distillation-based passage retrievers~\citep{izacard2020distilling,ren-etal-2021-rocketqav2}, which fine-tune the parameters of the retrievers directly on all training data by leveraging signals from cross-encoders.

\subsection{Pseudo Relevance Feedback}
Pseudo relevance feedback (PRF) techniques in information retrieval~\citep{rocchio1971relevance,lavrenko2001relevance} share a similar motivation to ours in that they refine query representations for a single test query.
Unlike using the true relevance feedback provided by users~\citep{baumgartner-etal-2022-incorporating}, PRF relies on heuristic or model-based relevance feedback, which can be easily automated.
Although most previous work uses PRF for sparse retrieval~\citep{croft2010search,zamani2018neural,li2018nprf,mao2020generation}, recent work has begun to apply PRF for dense retrieval~\citep{yu2021improving,wang2021pseudo,li2021pseudo}.

PRF aims to improve the quality of the retrieval by updating the initial query representation from the query encoder (i.e., $E_q(q) = \mathbf{q}_0$):
\begin{equation}
\begin{split}
    &\mathbf{q}_{t+1} \leftarrow g(\mathbf{q}_{t}, \topkret{q_t}),
\end{split}
\end{equation}
where $g$ is an update function and $\mathbf{q}_t$ denotes the query representation after $t$-th updates over $\qb_{0}$.

The classical Rocchio algorithm for PRF~\citep{rocchio1971relevance} updates the query representation as:
\begin{equation}\label{eq:prf-prev}
\begin{split}
    &g(\mathbf{q}_{t}, \topkret{q_t}) = \\ &\alpha \mathbf{q}_{t} + \beta\frac{1}{|\mathcal{C}_r|} \sum_{c_r \in \mathcal{C}_r} \mathbf{c}_r- \gamma\frac{1}{|\mathcal{C}_{nr}|} \sum_{c_{nr} \in \mathcal{C}_{nr}} \mathbf{c}_{nr},
\end{split}
\end{equation}
where $\mathcal{C}_{r}$ and $\mathcal{C}_{nr}$ denote \textit{relevant} and \textit{non-relevant} sets of contexts, respectively.
$\alpha$, $\beta$, and $\gamma$ determine the relative contribution of the current query representation $\qb_t$, relevant context representations $\cb_r$, and non-relevant context representations $\cb_{nr}$, respectively, when updating to $\qb_{t+1}$.
A common practice is to choose top-$k'$ contexts as pseudo-relevant among top-$k$ ($k' < k$), i.e., $\mathcal{C}_{r} = \mathcal{C}_{1:k'}^{q_t}$:
\begin{equation}\label{eq:prf}
\begin{split}
    &g(\mathbf{q}_{t}, \topkret{q_t}) = \\ 
    &\alpha \mathbf{q}_{t} + \beta\frac{1}{k'} \sum_{i=1}^{k'} \mathbf{c}_i- \gamma\frac{1}{k-k'} \sum_{i=k'+1}^k \mathbf{c}_{i}.
\end{split}
\end{equation}
\indent In this work, we theoretically show that our test-time query optimization is a generalization of the Rocchio algorithm.
While Eq.~\eqref{eq:prf} treats the positive (or negative) contexts equally, we use cross-encoder re-rankers~\citep{nogueira2019passage} to provide fine-grained pseudo labels and optimize the query representations with gradient descent.

\section{Methodology}

In this section, we provide an overview of our method (\Cref{sec:tqr}) and its two instantiations~(\Cref{sec:tqrmml}, \Cref{sec:variant}). 
We also introduce a relevance labeler for phrase retreival~(\Cref{sec:labeler}) and simple techniques to improve efficiency of \ours{} (\Cref{sec:inference}).

\subsection{Optimizing Test-time Query Representations}\label{sec:tqr}
We propose \ours~(Test-Time Optimization of Query Representations), which optimizes query representations at the instance level.
In our setting, the query and context encoders are fixed after training, and we optimize the query representations solely based on their retrieval results.
Figure~\ref{fig:overview} illustrates an overview of \ours.

First, given a single test query $q \in \mathcal{Q}_\text{test}$, we use a cross-encoder re-ranker $\cross{(\cdot)}$ to provide a score of how relevant each of the top-$k$ contexts $c \in \topkret{q}$ is with respect to a query:
\begin{equation}
    s = \cross{(q, c)},
\end{equation}
where $\cross{(\cdot)}$ is often parameterized with a pre-trained language model, which we detail in \Cref{sec:labeler}.
Compared to simply setting top-$k'$ results as pseudo-positive in PRF, using cross-encoders enables more fine-grained judgments of relevance over the top results.
In addition, it allows us to label results for \textit{test} queries as well without access to the gold label $c^*$.

\subsection{\ours{} with Hard Labels : \ours$_\text{hard}$}\label{sec:tqrmml}

First, we explore using the scores from the cross-encoder labeler~$\cross$ and selecting a set of pseudo-positive contexts $\toppret \subset \topkret{q}$ defined as the smallest set such that:
\begin{equation}\label{eq:topp}
\begin{split}
    P_k(\tilde{c}=&c^*|q, \cross) = \frac{\exp(\cross{(q, \tilde{c})}/\tau)}{\sum_{i=1}^k \exp(\cross{(q, c_i)}/\tau)} \\
    &\smashoperator{\sum_{\tilde{c}\in\toppret}} P_k(\tilde{c}=c^*|q, \cross) \geq p,
\end{split}
\end{equation}
where $\tau$ is a temperature parameter and $\tilde{c} \in \toppret$ denotes a pseudo-positive context selected by $\cross$.
Intuitively, we choose the smallest set of contexts as $\toppret$ whose marginal relevance with respect to a query under $\cross$ is larger than the threshold $p$.
This is similar to Nucleus Sampling for stochastic decoding~\citep{holtzman2019curious}.

Then, \ours{} optimizes the query representation with the gradient descent algorithm based on the relevance judgment $\toppret$ made by $\cross$:
\begin{equation}\label{eq:tql}
\begin{split}
    & \mathcal{L}
    _\text{hard}(q,\topkret{q}) = -\log\smashoperator{\sum_{\tilde{c}\in\toppret}} P_k(\tilde{c}|q),
\end{split}
\end{equation}
where $P_k(\tilde{c}|q)=\frac{\exp(\text{sim}(q, \tilde{c}))}{\sum_{i=1}^k \exp(\text{sim}(q, c_i))}$.
Similar to the query-side fine-tuning in Eq.~\eqref{eq:qsft}, we maximize the marginal likelihood of (pseudo) positive contexts $\toppret$.
We denote this version as \ours$_\text{hard}$.
Unlike query-side fine-tuning that updates the model parameters of $E_q(\cdot)$, we directly optimize the query representation $\mathbf{q}$ itself.
\ours$_\text{hard}$~is also an instance-level optimization over a single test query $q \in \mathcal{Q}_\text{test}$ without access to the gold label $c^*$.

For optimization, we use gradient descent:
\begin{equation}\label{eq:tql-sgd}
    \mathbf{q}_{t+1} \leftarrow \mathbf{q}_{t} - \eta \frac{\partial \mathcal{L}_\text{hard}(\mathbf{q}_{t}, \topkret{q_t})}{\partial \mathbf{q}_{t}},
\end{equation}
where $\eta$ denotes the learning rate for gradient descent and the initial query representation is used as $\mathbf{q}_0$.
Applying gradient descent over the test queries shares the motivation with dynamic evaluation for language modeling~\citep{krause2019dynamic}, but we treat each test query independently unlike the series of tokens for the evaluation corpus of language modeling.
For each iteration, we perform a single step of gradient descent followed by another retrieval with $\mathbf{q}_{t+1}$ to update $\mathcal{C}_{1:k}^{q_t}$ into $\mathcal{C}_{1:k}^{q_{t+1}}$.

\paragraph{{Relation to the Rocchio algorithm}}
Eq.~\eqref{eq:tql-sgd} could be viewed as performing PRF by setting the update function $g(\mathbf{q}_{t}, \topkret{q_t})=\mathbf{q}_{t} - \eta \frac{\partial \mathcal{L}_\text{hard}(\mathbf{q}_{t}, \topkret{q_t})}{\partial \mathbf{q}_{t}}$.
In fact, our update rule Eq.~\eqref{eq:tql-sgd} is a generalized version of the Rocchio algorithm as shown below:
\vspace{-0.1cm}
\begin{equation}\label{eq:generalized}
\begin{split}
  g(\mathbf{q}_{t}&, \topkret{q_t}) \\
  \quad=\mathbf{q}_t &+ \eta \sum_{\tilde{c}} \pctq(1 - \pkctq) \tilde{\mathbf{c}} \\
  \quad
  &-\eta \sum_{\tilde{c}} \big[ \pctq \smashoperator{\sum_{c \in \mathcal{C}_{1:k}^{q_t}, c\not=\tilde{c}}} \pkcq \mathbf{c} \big],
\end{split}
\end{equation}
\noindent where $\tilde{c} \in \mathcal{C}_\text{hard}^{q_t}$ and $\pctq=\frac{\exp(\text{sim}(q_t, \tilde{c}))}{\sum_{\tilde{c}'} \exp(\text{sim}(q_t, \tilde{c}'))}$ (proof in \Cref{sec:apdx-mml}).
Although our update rule seems to fix $\alpha$ in Rocchio to $1$, it can be dynamically changed by applying weight decay during gradient descent, which sets $\alpha = 1-\eta\lambda_\text{decay}$ multiplied by $\qb_t$.
Then, the equality between Eq.~\eqref{eq:prf} and Eq.~\eqref{eq:generalized} holds when $\mathcal{C}_\text{hard}^{q_t} = \mathcal{C}_{1:k'}^{q_t}$ with $P_k(c|q_t)$ being equal for all $c \in \mathcal{C}_{1:k}^{q_t}$, namely $P_k(c|q_t)=1/k$.
This reflects that the Rocchio algorithm treats all top-$k'$ results equally (i.e., $P(\tilde{c}|q_t) = 1/k'$).
Then, $\beta=\gamma=\eta\frac{k-k'}{k}$ holds (\Cref{sec:apdx-equality}).

In practice, $\mathcal{C}_\text{hard}^{q_t}$ would be different from $\mathcal{C}_{1:k'}^{q_t}$ if some re-ranking happens by $\cross$.
Also, each pseudo-positive context vector $\tilde{\mathbf{c}}$ in the second term of the RHS of Eq.~\eqref{eq:generalized} has a different weight.
The contribution of $\tilde{\mathbf{c}}$ is maximized when it has a larger probability mass $\pctq$ among the pseudo-positive contexts, but a smaller probability mass $\pkctq$ among the top-$k$ contexts; this is desirable since we want to update $\mathbf{q}_t$ a lot when the initial ranking of pseudo-positive context in top-$k$ is low.
For instance, if there is a single pseudo-positive context $\tilde{c}$ (i.e., $\pctq) = 1$) ranked at the bottom of top-$k$ with a large margin with top-1 (i.e., $\pkctq = 0$), then $\pctq(1-\pkctq)=1$ is maximized.

\subsection{\ours{} with Soft Labels : \ours$_\text{soft}$}\label{sec:variant}

From Eq.~\eqref{eq:generalized}, we observe that it uses pseudo-positive contexts $\mathcal{C}_\text{hard}^{q_t}$ sampled by the cross-encoder labeler $\cross$, but the contribution of $\tilde{\mathbf{c}}$ (the second term in RHS) does not directly depend on the scores from $\cross$.
The scores are only used to make a hard decision in pseudo-positive contexts.
Another version of \ours{} uses the normalized scores of a cross-encoder over the retrieved results as soft labels.
We can simply change the maximum marginal likelihood objective in Eq.~\eqref{eq:tql} to reflect the scores from $\cross$ in~$g$.
Specifically, we change Eq.~\eqref{eq:tql} to minimize Kullback-Leibler (KL) divergence loss as follows:
\begin{equation}\label{eq:tql-kl}
\begin{split}
    &\mathcal{L}_\text{soft}(\mathbf{q}_{t},\mathcal{C}_{1:k}^{q_t}) = \\ &\quad\quad-\smashoperator{\sum_{i=1}^k} P(c_i|q_t,\cross) \log \frac{P_k(c_i|q_t)}{P(c_i|q_t, \cross)},
\end{split}
\end{equation}
where $P(c_i|q_t, \cross)=P(c_i=c^*|q_t, \cross)$ defined in Eq.~\eqref{eq:topp}. 
We call this version \ours$_\text{soft}$.
The update rule $g$ for \ours$_\text{soft}$~changes as follows:
\begin{equation}\label{eq:generalized-kl}
\begin{split}
    &~g(\mathbf{q}_{t}, \mathcal{C}_{1:k}^{q_t}) \\
    &=\mathbf{q}_t + \eta\sum_{i=1}^k P(c_i|q_t, \cross) \mathbf{c}_i - \eta\sum_{i=1}^k P_k(c_i|q_t)\mathbf{c}_i.
\end{split}
\end{equation}
Eq.~\eqref{eq:generalized-kl} shows that $\mathbf{q}_{t+1}$ reflects $\mathbf{c}_i$ weight-averaged by the cross-encoder (i.e., $P(c_i|q_t,\cross)$) while removing $\mathbf{c}_i$ weight-averaged by the current retrieval result (i.e., $P_k(c_i|q_t)$)~(proof in \Cref{sec:apdx-kl}).

\subsection{{Relevance Labeler for Phrase Retrieval}}\label{sec:labeler}
{In the previous section, we used a cross-encoder reranker $\cross$ to provide a relevance score $s_i$ over a pair of a query $q$ and a context $c$.}
While it is possible to use an off-the-shelf re-ranker~\citep{fajcik2021r2} for passage retrieval, no prior work has introduced a re-ranker for phrase retrieval~\cite{lee2021phrase}.
In this section, we introduce a simple and accurate phrase re-ranker for \ours.

\paragraph{{Inputs for re-rankers}}
For phrase retrieval, sentences containing each retrieved phrase are considered as contexts, following~\citet{lee2021phrase}.
For each context, we also prepend the title of its document and use it as our context for re-rankers.
To train our re-rankers, we first construct a training set from the retrieved contexts of the phrase retriever given a set of training queries $\mathcal{Q}_\text{train}$.
Specifically, from the top retrieved contexts $\mathcal{C}_{1:k}$ for every $q \in \mathcal{Q}_\text{train}$, we sample one positive context $c_q^+$ and one negative context $c_q^-$.
In open domain QA, it is assumed that a context that contains a correct answer to each $q$ is relevant (positive).
Our re-ranker is trained on a dataset $\mathcal{D}_\text{train} = \{(q, c_q^+, c_q^-) | q \in \mathcal{Q}_\text{train}\}$.

\paragraph{Architecture}
We use the RoBERTa-large model~\citep{liu2019roberta} as the base model for our re-ranker.
Given a pre-trained LM $\mathcal{M}$, the cross-encoder re-ranker $\cross$ outputs a score of a context being relevant:
\begin{equation}
\begin{split}
    s &= \cross{(q, c)} = \wb^\top \mathcal{M}(q \textsc{ $\oplus$}~c) \textsc{[CLS]}
\end{split}
\end{equation}
where $\{\mathcal{M}, \mathbf{w}\}$ are the trainable parameters and $\oplus$ denotes a concatenation of $q$ and $c$ using a [SEP] token.
Since phrase retrievers return both phrases and their contexts, we use special tokens [S] and [E] to mark the retrieved phrases within the contexts.

Re-rankers are trained to maximize the probability of a positive context $c_q^+$ for every $(q, c_q^+, c_q^-) \in \mathcal{D}_\text{train}$.
We use the binary cross-entropy loss defined over the probability $P^+ = \frac{\exp(\mathbf{h}^+)}{\exp(\mathbf{h}^+) + \exp(\mathbf{h}^-)}$ where $\mathbf{h}^+ = \cross(q, c_q^+)$ and $\mathbf{h}^- = \cross(q, c_q^-)$.
{We pre-train $\cross$ on reading comprehension datasets~\citep{rajpurkar2016squad,joshi2017triviaqa,kwiatkowski2019natural}, which helped improve the quality of $\cross$.
For the ablation study of our phrase re-rankers, see~\Cref{sec:apdx-detail} for details.}

\paragraph{Score aggregation}
{After running \ours, aggregating the reranking scores with the retreival scores provides consistent improvement.}
Specifically, we linearly interpolate the similarity score $\text{sim}(q, c_i)$ with the re-ranking score $s_i$ and use this to obtain the final results: $\lambda s_i + (1-\lambda) \text{sim}(q, c_i)$.

\subsection{{Efficient Implementation of \ours}}\label{sec:inference}
\ours{} aims to improve the recall of gold candidates by iteratively searching with updated query representations.
However, it has high computational complexity, since it needs to label top-$k$ retrieval results with a cross-encoder and perform additional retrieval.
To minimize the additional time complexity, we perform up to $t$ = 3 iterations with early stopping conditions.
Specifically, at every iteration of \ours$_\text{hard}$, we stop when the top-1 retrieval result is pseudo-positive, i.e., $c_1 \in \mathcal{C}_\text{hard}^{q_t}$.
When using \ours$_\text{soft}$, we stop iterating when the top-1 retrieval result has the highest relevance score.
Additionally, we cache $\cross{(q, c_i)}$ for each query to skip redundant computation.

\section{Experiments}

\begin{table*}[t!]
    \centering
    \resizebox{2\columnwidth}{!}{%
    \begin{tabular}{lcclllll}
    \toprule
    \textbf{Model} & Top-$k$ & s/q ($\downarrow$) & \textsc{NQ} & \textsc{TriviaQA} & \textsc{WQ} & \textsc{TREC} & \textsc{SQuAD}\\
    \midrule
    \textit{Retriever + Extractive Reader}    \\
    \midrule
    DPR$_\text{multi}$~\citep{karpukhin2020dense} & & & 41.5 & 56.8 & 42.4 & 49.4 & 24.1  \\
    {\quad + Re-ranker}~\citep{iyer2021reconsider} & 5 & 1.21 &  43.1 \ua{1.6} & 59.3 \ua{2.5}& 44.4 \ua{2.0} & 49.3 \da{0.1} & - \\ 
    GAR~\citep{mao2020generation} & & & 41.8 & 62.7 & - & - & -\\
    DPR$_\text{multi}$ (large) &  &  &44.6 & 60.9 & 44.8 & 53.5 & -\\ 
    \quad + Re-ranker & 5  & {>1.21$^{*}$} &  45.5 \ua{0.9} & 61.7 \ua{0.8}& \textbf{45.9 \mua{1.1}} & \textbf{55.3 \mua{1.8}} & - \\ 
    ColBERT-QA$_\text{large}$~\citep{khattab2021relevance} & & & 47.8 & \textbf{70.1} & - & - & \textbf{54.7} \\ 
    UnitedQA-E$_\text{large}$ & & & \textbf{51.8} & 68.9 & - & - & - \\
    \midrule
    \textit{Retriever-only}  \\

    \midrule
    DensePhrases$_\text{multi}$~\citep{lee2021learning} & & & 41.6 & 56.3 & 41.5 & 53.9 & 34.5\\
    \quad + PRF$_\text{Rocchio}$ & 10 & 0.09 & 41.6 \na{0.0}& 56.5 \ua{0.2}& 41.7 \ua{0.2}& 54.0 \ua{0.1}& 34.9 \ua{0.4}\\
    \quad + Phrase re-ranker ({Ours}) &10 & 0.24 &  47.0 \ua{5.4} & 65.4 \ua{9.1}& 45.9 \ua{4.4} & 60.5 \ua{6.6}& 43.1 \ua{8.6}\\
    \quad + Phrase re-ranker ({Ours}) &40&  1.04 & 46.5 \ua{4.9} & 66.0 \ua{9.7} & 46.3 \ua{4.8} & 61.5 \ua{7.6} & 45.3 \ua{10.8}\\ 
    \quad + \ours$_\text{hard}$ ({Ours}) &10& 0.44 & \textbf{48.6 \mua{7.0}} & 66.4 \ua{10.1} & 46.1 \ua{4.6} & 62.0 \ua{8.1}& 45.2 \ua{10.7}\\
    \quad + \ours$_\text{hard}$ ({Ours}) &20 & 0.78 &47.9 \ua{6.3}& \textbf{66.8} \textbf{\mua{10.5}}& \textbf{46.9} \textbf{\mua{5.4}}& 62.5 \ua{8.6}& \textbf{46.4} \textbf{\mua{11.9}}\\
    \quad + \ours$_\text{soft}$ ({Ours}) &10& 0.43 & 47.9 \ua{6.3}&66.5 \ua{10.2}&46.3 \ua{4.8}&\textbf{63.1 \mua{9.2}}&44.9 \ua{10.4}\\
     \quad + \ours$_\text{soft}$ ({Ours}) &20 & 0.78 & 47.6 \ua{6.0}& 66.6 \ua{10.3}& \textbf{46.9 \mua{5.4}}& 62.5 \ua{8.6}& 46.0 \ua{11.5}\\
    \bottomrule
    \end{tabular}
    }
    \caption{Open-domain QA results. We report exact match (EM) on each test set.
    {s/q denotes the average latency of a single query in seconds, which includes the latency of {DPR$_\text{multi}$ or} DensePhrases$_\text{multi}$.} 
    {For re-ranking and PRF-based methods, we denote the improvement from their base retrievers in \ua{x.x}}.
    Best performance is denoted in \textbf{bold}. 
    {*: We could not measure the exact latency of DPR$_\text{multi}$ (large) as its checkpoint has not been released.}
    }
    \vspace{+0.1cm}
    \label{tab:od-qa}
\end{table*}

We test \ours{} on multiple open-domain QA datasets.
Specifically, we evaluate its performance on phrase retrieval and passage retrieval.

\subsection{Datasets}\label{sec:dataset}
{We mainly use six open-domain QA datasets: Natural Questions~\citep{kwiatkowski2019natural}, TriviaQA~\citep{joshi2017triviaqa}, WebQuestions~\citep{berant2013semantic}, CuratedTrec~\citep{baudivs2015modeling}, SQuAD~\citep{rajpurkar2016squad}, and EntityQuestions~\citep{sciavolino2021simple}.}
Following previous works, Entity Questions is only used for testing.
See statistics in \Cref{sec:apdx-datastats}.

\subsection{Open-domain Question Answering}\label{sec:exp-odqa}
For end-to-end open-domain QA, we use phrase retrieval~\citep{seo2019real,lee2021learning} for \ours, which directly retrieves phrases from the entire Wikipedia using a phrase index.
Since a single-stage retrieval is the only component in phrase retrieval, it is easy to show how its open-domain QA performance can be directly improved with \ours.
We use DensePhrases~\citep{lee2021learning} for our base phrase retrieval model and train a cross-encoder labeler as described in~\Cref{sec:labeler}.
{We report exact match~(EM) for end-to-end open-domain QA.}
We use $k=\{10, 40\}$ for our phrase re-ranker and $k=\{10, 20\}$ for \ours{} on open-domain QA while $k=10$ is used for both whenever it is omitted. 
{}
For the implementation details of \ours, see \Cref{sec:apdx-detail}.

\paragraph{Baselines}
Many open-domain QA models take the retriever-reader approach~\citep{chen2017reading,lee2019latent,izacard2020leveraging,singh2021end}.
As our baselines, we report extractive open-domain QA models, which is a fair comparison with retriever-only (+ re-ranker) models whose answers are always extractive.
For re-ranking baselines of retriever-reader models, we report ReConsider~\citep{iyer2021reconsider}, which re-ranks the outputs of DPR + BERT.
For a PRF baseline, GAR~\citep{mao2020generation}, which uses context generation models for augmenting queries in BM25, is reported.

\paragraph{Results}
Table~\ref{tab:od-qa} shows the results on the five open-domain QA datasets in the in-domain evaluation setting where all models use the training sets of each dataset they are evaluated on.
First, we observe that using our phrase re-ranker largely improves the performance of DensePhrases$_\text{multi}$.
{Compared to adding a re-ranker on the retriever-reader model (DPR$_\text{multi}$ + Re-ranker by \citealp{iyer2021reconsider}), {our phrase re-ranking approach performs 5$\times$ faster with a larger top-$k$ due to the efficient retriever-only method. Furthermore,} the performance gain is significantly larger possibly due to the high top-$k$ accuracy of phrase retrievers.}
{Unlike using the Rocchio algorithm, using \ours$_\text{hard}$ or \ours$_\text{soft}$ greatly improves the performance of the base retriever.}
{Compared to our phrase re-ranker$_{k=40}$, \ours$_{\text{hard},{k=20}}$ runs 1.3$\times$ faster as well as outperforming it by up to 2.0\%.}
Even \ours$_{\text{hard},{k=10}}$ often outperforms re-ranker$_{k=40}$ with 2.4$\times$ faster inference.
{For this task, \ours$_\text{hard}$ and \ours$_\text{soft}$ work similarly with exceptions on NQ and TREC.}

\begin{figure}[t]
\centering
\includegraphics[width=0.8\columnwidth]{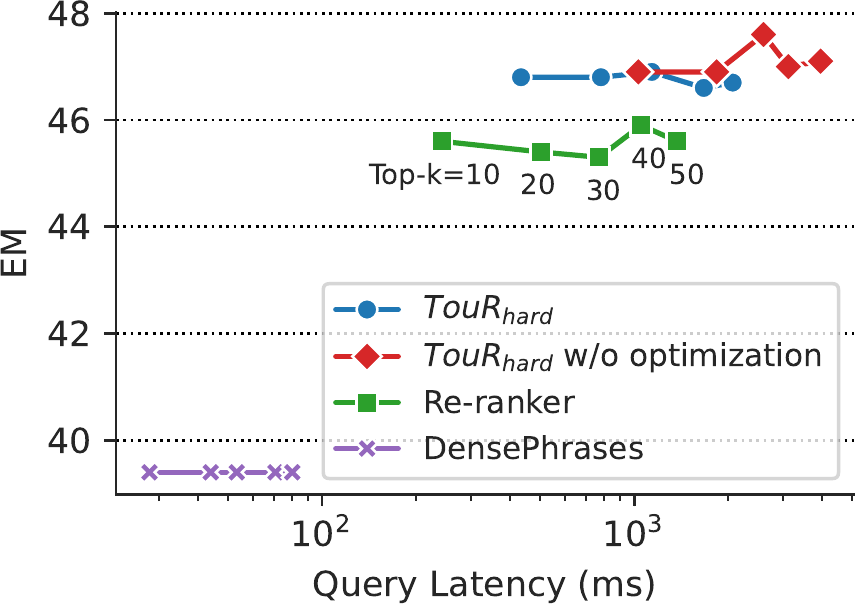}
\caption{
{
Query latency (ms) vs. open-domain QA performance (EM) of different models on the NQ development set.
Query latency is controlled by varying the top-$k$ value incrementally.
\ours$_\text{hard}$~w/o optimization: \ours$_\text{hard}$~without the efficient implementation in~\Cref{sec:inference}.
} 
}
\vspace{+0.1cm}
\label{fig:time_comparison}
\end{figure}

\begin{table*}[t!]
    \centering
    \resizebox{2.1\columnwidth}{!}{%
    \begin{tabular}{lcclllllll}
    \toprule
    & && \textit{Training} &&\multicolumn{5}{c}{\textit{Unseen Query Distribution}}\\
    \cmidrule{4-4}\cmidrule{6-10}
    \textbf{Model} & Top-$k$ & s/q ($\downarrow$) & \textsc{NQ} && \textsc{TriviaQA} & \textsc{WQ} & \textsc{TREC} & \textsc{Squad} & \textsc{EntityQ}\tablefootnote{{While the passage retrieval accuracy is mostly reported for Entity Questions~\citep{ram-etal-2022-learning, lewis-etal-2022-boosted}, we also report EM for open-domain QA.}}\\

    \midrule
    DPR$_\text{NQ}$$^*$~\citep{karpukhin2020dense} &&& 39.4 && 29.4 & - & - & 0.1 & -\\ %
    DensePhrases$_\text{NQ}$~\citep{lee2021learning} &&& 40.8 && 33.4 & 23.8 & 33.6 & 15.4 & 22.4\\
    \quad + Phrase re-ranker~({Ours})&10&0.24& 45.4 \ua{4.6} && 40.9 \ua{7.5} & 26.6 \ua{2.8} & 37.8 \ua{4.2} & 20.2 \ua{4.8} & 26.8 \ua{4.4}\\
    \quad + Phrase re-ranker~({Ours})&40&1.04& 44.5 \ua{3.7} &&  41.7 \ua{8.3} & 26.4 \ua{2.6} & 37.8 \ua{4.2} & 21.4 \ua{6.0} & 27.1 \ua{4.7}\\
    \quad + \ours$_\text{hard}$~{(Ours)} &10&0.44& \textbf{47.0 \mua{6.2}} && 42.6 \ua{9.2} & 27.7 \ua{3.9} & 38.3 \ua{4.7} & 21.5 \ua{6.1} & 27.9 \ua{5.5} \\
    \quad + \ours$_\text{hard}$~{(Ours)}&20&0.78& 46.5 \ua{5.7} &&  \textbf{42.9 \mua{9.5}} & \textbf{28.2 \mua{4.4}} & \textbf{39.8 \mua{6.2}} & \textbf{22.1 \mua{6.7}} & \textbf{28.3 \mua{5.9}}\\
    \quad + \ours$_\text{soft}$~{(Ours)} &10&0.43& 46.2 \ua{5.4} && 42.5 \ua{9.1}& 27.4 \ua{3.6}& 38.2 \ua{4.6} & 21.2 \ua{5.8} & 27.6 \ua{5.2}\\
    \quad + \ours$_\text{soft}$~{(Ours)} &20&0.78& 45.7 \ua{4.9} &&  42.7 \ua{9.3} & 27.7 \ua{3.9} & \textbf{39.8 \mua{6.2}} & 21.7 \ua{6.3} & 27.9 \ua{5.5}\\
    \bottomrule
    \end{tabular}
    }
    \caption{
    Open-domain QA results under query distribution shift.
    All retrievers and re-rankers are trained on NaturalQuestions and evaluated on unseen query distributions.
    EM is reported on each test set. 
    $^*$: results obtained from the official implementation, which does not support running end-to-end QA on WebQuestions, CuratedTREC, and EntityQuestions.
    \ua{x.x} shows EM improvement from DensePhrases$_\text{NQ}$.
    }
    \vspace{+0.1cm}
    \label{tab:od-qa-ood}
\end{table*}

\begin{table*}[t]
    \centering
    \resizebox{0.8\textwidth}{!}{%
    \begin{tabular}{llllllll}
        \toprule
         &\multicolumn{1}{c}{\textsc{NQ}} & \multicolumn{1}{c}{\textsc{TriviaQA}} && \multicolumn{1}{c}{\textsc{EntityQ$^\dagger$}} \\
        \textbf{Model} & \multicolumn{1}{c}{(Acc@20/100)} & \multicolumn{1}{c}{(Acc@20/100)} & & \multicolumn{1}{c}{(Acc@20/100)}\\

        \midrule
        DensePhrases$_\text{multi}$ & 79.8 / 86.0 & 81.6 / 85.8 & & 61.0 / 71.2\\
        \quad + Re-ranker~\citep{fajcik2021r2} & 83.2 / 86.0 \ua{3.4} & 83.0 / 85.8 \ua{1.4} & & 65.3 / 71.2 \ua{4.3}\\
        \quad + \ours$_\text{hard}$~{(Ours)}& 84.0 / 86.9 \ua{4.2} & \textbf{83.2} / \textbf{86.1} \textbf{\mua{1.6}}& & \textbf{66.2} / \textbf{72.4} \textbf{\mua{5.2}}\\
        \quad + \ours$_\text{soft}$~{(Ours)}&  \textbf{84.2 / 87.0 \mua{4.4}} & \textbf{83.2} / \textbf{86.1} \textbf{\mua{1.6}} & & \textbf{66.2 / 72.4} \textbf{\mua{5.2}}\\

        \midrule
        DPR$_\text{multi}$ & 79.4 / 86.5 & 79.0 / 84.8 & & 57.9 / 70.8\\
        \quad + Re-ranker~\citep{fajcik2021r2} & 83.6 / 86.5 \ua{4.2} & \textbf{81.6} / 84.8 \mua{\textbf{2.6}} & & 64.4 / 70.8 \ua{6.5}\\
        \quad + \ours$_\text{hard}$~{(Ours)}& 84.0 / 87.0 \ua{4.6} & 81.5 / 84.9 \ua{2.5} & & 65.6 / 71.9 \ua{7.7}\\
        \quad + \ours$_\text{soft}$~{(Ours)}& \textbf{84.2 / 87.2 \mua{4.8}} &  \textbf{81.6} / \textbf{85.1} \mua{\textbf{2.6}}& & \textbf{66.2} / \textbf{72.5} \textbf{\mua{8.3}}\\
        \bottomrule
    \end{tabular}
    }
    \caption{Passage retrieval results. We report  Acc@20 / Acc@100 (\%) on each test set. Each retriever (and re-ranker) is trained on multiple open-domain QA datasets described in~\Cref{sec:dataset}, which makes Natural Questions and TriviaQA in-domain evaluation and leaves EntityQuestions as out-of-domain evaluation. 
    We denote improvement in Acc@20 from DensePhrases$_\text{multi}$ or DPR$_\text{multi}$ in \ua{x.x}.
    {$^\dagger$: unseen query distribution.}
    }
    \vspace{+0.1cm}
    \label{tab:od-qa-pr}
\end{table*}

\paragraph{{Latency vs. performance}}\label{sec:runtime}
{Figure~\ref{fig:time_comparison} compares the query latency and performance of \ours{} and other baselines on the NQ development set.
We vary the top-$k$ value from 10 to 50 by 10 (left to right) to visualize the trade-off between latency and performance.
The result shows that \ours{} with only top-$10$ is better and faster than the re-ranker with the best top-$k$.
Specifically, \ours$_{\text{hard},k=10}$ outperforms re-ranker$_{k=40}$ by 1.0\% while being 2.5$\times$ faster.
This shows that \ours{} requires a less number of retrieval results to perform well, compared to a re-ranker model that often requires a larger $k$.
}

\paragraph{{Query distribution shift}}
{In Table~\ref{tab:od-qa-ood}, we show open-domain QA results under query distribution shift from the training distribution.}
{Compared to DensePhrases$_\text{multi}$ in \Cref{tab:od-qa}, which was trained on all five open-domain QA datasets, we observe huge performance drops on unseen query distributions when using DPR$_\text{NQ}$ and DensePhrases$_\text{NQ}$.}
DPR$_\text{NQ}$ seems to suffer more (e.g., 0.1 on SQuAD) since both of its retriever and reader were trained on NQ, which exacerbates the problem when combined.

On the other hand, using \ours{} largely improves the performance of DensePhrases$_\text{NQ}$ on many unseen query distributions even though all of its component were still trained on NQ.
{Specifically, \ours$_{\text{hard},k=20}$ gives 6.5\% improvement on average across different query distributions, which easily outperforms our phrase re-ranker$_{k=40}$.}
Interestingly, \ours$_\text{hard}$ consistently performs better than \ours$_\text{soft}$ in this setting, which requires more investigation in the future.

\subsection{Passage Retrieval}
We test \ours{} on the passage retrieval task for open-domain QA.
We use DPR as a passage retriever and DensePhrases as a phrase-based passage retriever~\citep{lee2021phrase}.
In this experiment, we use an off-the-shelf passage re-ranker~\citep{fajcik2021r2} to show how existing re-rankers can serve as a pseudo labeler for \ours.
We report the top-$k$ retrieval accuracy, which is 1 when the answers exist in top-$k$ retrieval results.
For passage retrieval, we use $k=100$ for both the re-ranker and \ours{} due to the limited resource budget.

\paragraph{Results}
Table~\ref{tab:od-qa-pr} shows the results of passage retrieval for open-domain QA.
{We find that using \ours{} consistently improves the passage retrieval accuracy.}
{Under the query distribution shift similar to~\Cref{tab:od-qa-ood}, DPR$_\text{multi}$ + \ours$_\text{soft}$ improves the original DPR by 8.3\% and advances the off-the-shelf re-ranker by 1.8\% on EntityQuestions (Acc@20).}
Notably, Acc@100 always improves with \ours, which is not possible for re-rankers since they do not update the top retrieval results.
{Unlike the phrase retrieval task, we observe that \ours$_\text{soft}$~is a slightly better option than \ours$_\text{hard}$~on this task.}

\section{Analysis}

\begin{table}[t]
    \centering
    \resizebox{1\columnwidth}{!}{%
    \begin{tabular}{lccccc}
    \toprule 
    & & \multicolumn{3}{c}{\textit{Overlap}} \\
    \cmidrule{3-5}
    \textsc{NQ} & Total & Query & Answer$_\text{only}$ & None \\
    \midrule
DensePhrases$_\text{multi}$ & 41.3 & 63.3 & 33.7 & 23.9 \\
Re-ranker~({Ours}) & 46.8 & 66.7 & 39.0 & 31.0 \\
\ours$_\text{hard}$~({Ours}) & \textbf{48.6} & \textbf{70.1} & \textbf{40.3} & \textbf{33.7} \\
\midrule
\textsc{TriviaQA} \\ 
\midrule
DensePhrases$_\text{multi}$ & 53.8 & 76.5 & 46.2 & 32.6 \\
Re-ranker~({Ours}) & 62.8 & 82.1 & 60.3 & 41.5 \\
\ours$_\text{hard}$~({Ours}) & \textbf{63.8} & \textbf{83.6} & \textbf{62.3} & \textbf{42.2} \\
\midrule
    \textsc{WQ} \\ 
    \midrule
DensePhrases$_\text{multi}$ & 41.5 & 70.8 & 39.5 & 27.5 \\
Re-ranker~({Ours}) & 45.9 & \textbf{73.4} & \textbf{48.5} & 31.5 \\
\ours$_\text{hard}$~({Ours})  & \textbf{46.2} & 70.1 & \textbf{48.5} & \textbf{33.0} \\
    \bottomrule
    \end{tabular}
    }
    \caption{
    {Open-domain QA results on train-test overlap splits by~\citet{lewis2020question}.
    \textbf{Query} overlap denotes test queries that are paraphrases of training queries. 
    \textbf{Answer$_\text{only}$} overlap denotes test queries that have answers present in training data, while their queries are not overlapping with any training queries. \textbf{None} overlap denotes test queries without any query or answer overlap with training data.
    We report EM on each split.}
    }
    \label{tab:qa-overlap}
    \vspace{+0.1cm}
\end{table}

\begin{table}[t]
    \centering
    \resizebox{0.8\columnwidth}{!}{%
    \begin{tabular}{lc}
        \toprule
        &\textsc{NQ} \\
        \midrule
        DensePhrases$_\text{NQ}$ & 42.4 \\
        \midrule
        DensePhrases$_\text{NQ}$ + \ours$_\text{hard}$&  \textbf{48.4} \\
        \quad $\toppret \Rightarrow \mathcal{C}_{1:k'}$ ($k' = 3$) & 46.1 \\
        \quad SGD $\Rightarrow$ interpolation ($\beta=0.3$) & 48.2 \\
        \quad $\lambda=0.1$ $\Rightarrow$ $\lambda = 0$ & 48.1 \\
        \quad $\lambda=0.1$ $\Rightarrow$ $\lambda = 1$ & 48.0 \\
        \quad $\ours_\text{hard} \Rightarrow \ours_\text{soft}$ & 47.7 \\
        \bottomrule
    \end{tabular}
    }
    \caption{
    Ablation study of \ours$_\text{hard}$~on 
the Natural Questions (\textsc{NQ}) development set.
    We report EM for end-to-end open-domain QA.
    }
    \vspace{+0.1cm}
    \label{tab:ablation}
\end{table}

\subsection{{Train-Test Overlap Analysis}}

Open-domain QA datasets often contain semantically overlapping queries and answers between training and test sets~\citep{lewis2020question}, which overestimates the generalizability of QA models.
Hence, we test our models on train-test overlap splits provided by~\citet{lewis2020question}.
\Cref{tab:qa-overlap} shows that \ours{} consistently improves the performance of test queries that do not overlap with training data (i.e., None).
Notably, on WebQuestions, while the performance on the none overlap split has been improved by 1.5\% from the re-ranker, the performance on query overlap is worse than the re-ranker since unnecessary exploration is often performed on overlapping queries.
{Our finding on the effectiveness of query optimization is similar to that of~\citet{mao2020generation}, while our approach often improves performance on query overlap cases.}

\begin{figure}[t]
\centering
\includegraphics[width=0.8\columnwidth]{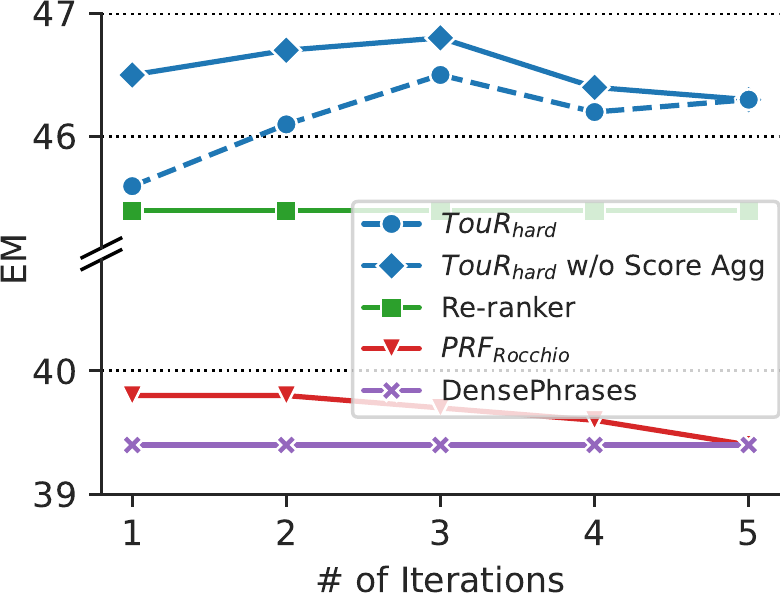}
\caption{
{Effect of multiple iterations in~\ours$_\text{hard}$ and PRF$_\text{Rocchio}$}. 
{We report open-domain QA EM on the Natural Questions development set.}
{Score Agg: score aggregation between the re-ranker and the retriever.}
{Note that the performance of original DensePhrases and Re-ranker is not affected by iterations.
}
}
\vspace{+0.1cm}
\label{fig:iteration}
\end{figure}

\begin{figure}[t!]
\centering
\includegraphics[width=0.8\columnwidth]{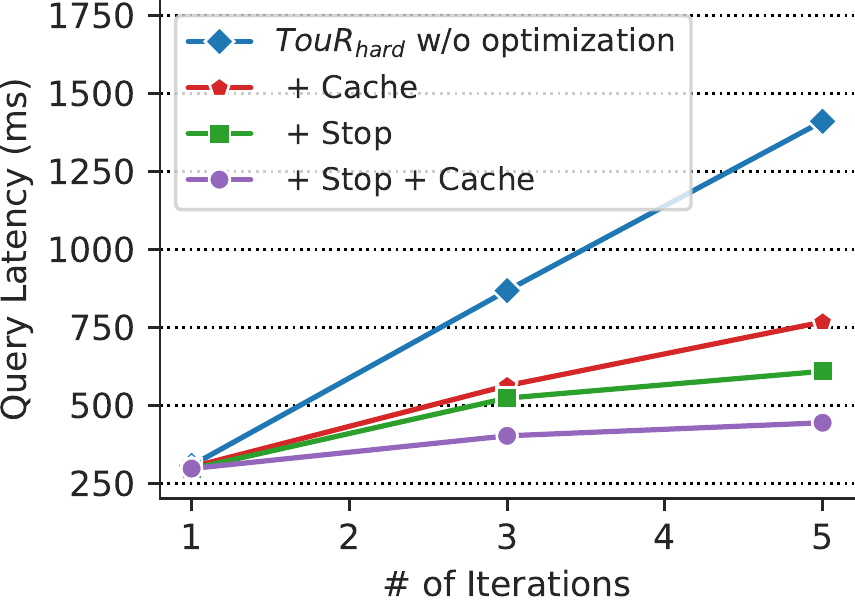}
\caption{
{Ablation study of efficient implementation of~\ours$_\text{hard}$. Latency is reported for different numbers of iterations. Cache: caching $\cross(q, c_i)$ for every iteration. Stop: applying the stop condition of $c_1 \in \mathcal{C}_\text{hard}^{q_t}$.
}}
\vspace{+0.1cm}
\label{fig:time_complexity}
\end{figure}

\begin{figure*}[t]
\centering
\includegraphics[width=2\columnwidth]{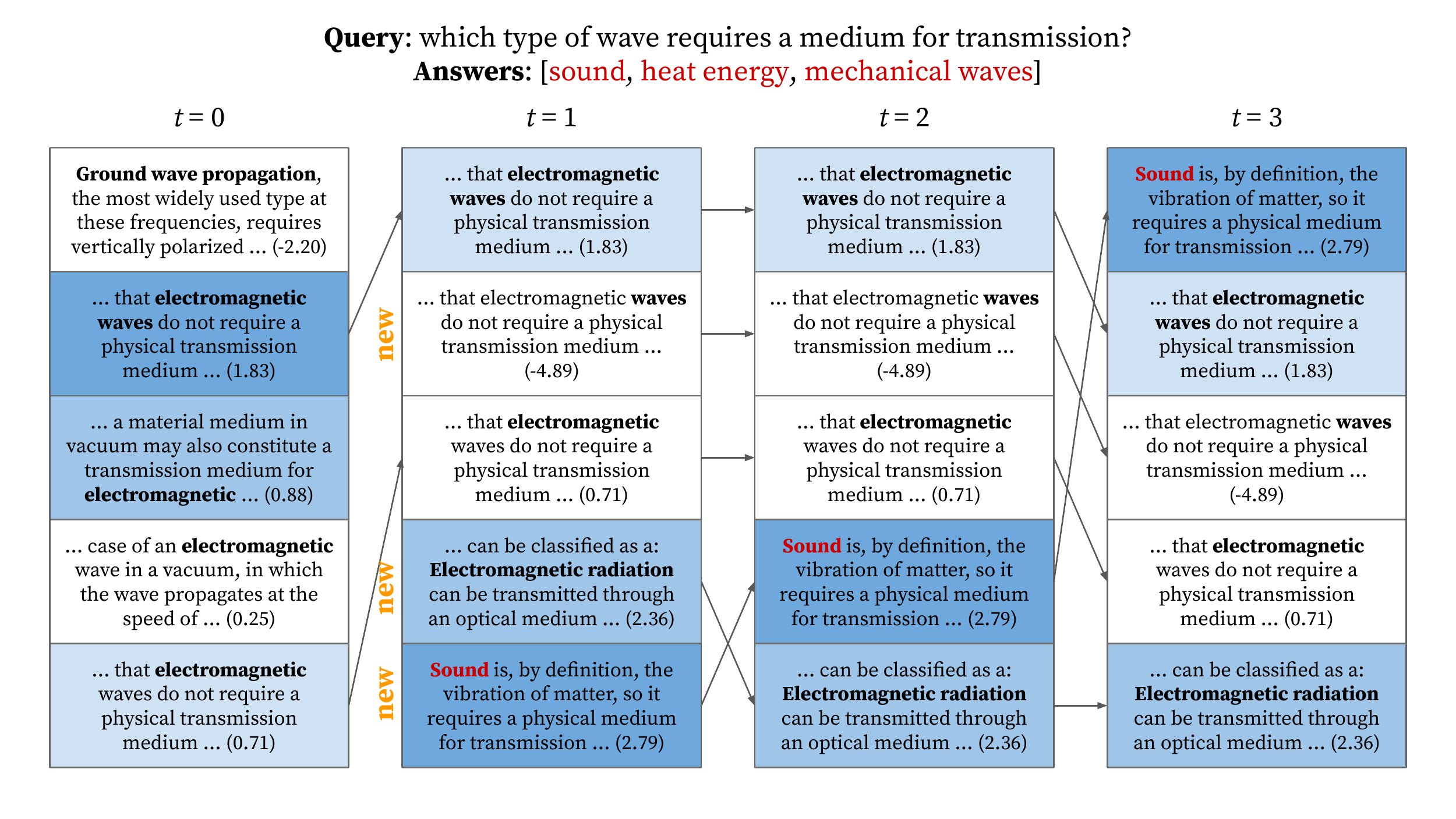}
\caption{
A sample prediction of \ours$_\text{hard}$ from Natural Questions. For every $t$-th iteration of \ours$_\text{hard}$, we show the top 5 phrases (denoted in bold) retrieved from DensePhrases along with their passages.
The score $s_i$ from the cross-encoder labeler $\cross$ is shown in each parenthesis.
$t=0$ denotes initial retrieval results. When $t=1$, \ours$_\text{hard}$ obtains three new results and the correct answer ``\textcolor{red}{Sound}'' becomes the top-1 prediction at $t=3$.
}
\label{fig:examples}
\end{figure*}

\subsection{{Ablation Study}}\label{sec:ablation}

{Table~\ref{tab:ablation} shows an ablation study of \ours$_\text{hard}$~on end-to-end open-domain QA.}
{
We observe that using fine-grained relevance signals generated by our phrase re-ranker (i.e., $\toppret$) is significantly more effective than simply choosing top-$k'$ as relevance signals (i.e., $\mathcal{C}_{1:k'}$).
Using SGD or aggregating the final scores between the retriever and the re-ranker gives additional improvement.}

{Figure~\ref{fig:iteration}~shows the effect of multiple iterations in \ours$_\text{hard}$~compared to the Rocchio algorithm.}
{While $\text{PRF}_\text{Rocchio}$ with $t=1$ achieves slightly better performance than DensePhrases, it shows a diminishing gain with a larger number of iterations.}
{In contrast, the performance of \ours$_\text{hard}$~benefits from multiple iterations until $t=3$.}
{Removing the score aggregation between the retriever and the re-ranker~(i.e., $\lambda=0$) causes a performance drop, but it quickly recovers with a larger $t$.}

\paragraph{{Efficient implementation}}

{Simple techniques introduced in~\Cref{sec:inference} such as early stopping and caching significantly reduce the run-time of~\ours.
{Figure~\ref{fig:time_complexity} summarizes the effect of optimization techniques to improve efficiency of~\ours.
Without each technique, the latency increases linearly with the number of iterations.
By adding the caching mechanism for $\cross$ and the stop condition of $c_1 \in \mathcal{C}_\text{hard}^{q_t}$, the latency is greatly reduced.}

\paragraph{{Prediction sample}}
Figure~\ref{fig:examples} shows a sample prediction of~\ours.
We use DensePhrases$_\text{multi}$ + \ours$_\text{hard}$~with $k=10$, from which the top-5 results are shown.
While the initial result at $t=0$ failed to retrieve correct answers in the top-10, the next round of \ours$_\text{hard}$ gives new results including the correct answer, which were not retrieved before.
{As the iteration continues, the correct answer starts to appear in the top retrieval results, and becomes the top-1 at $t=3$.}

\section{Conclusion}

In this paper, we propose \ours, which iteratively optimizes test query representations for dense retrieval.
Specifically, we optimize instance-level query representations at test time using the gradient-based optimization method over the top retrieval results.
We use cross-encoder re-rankers to provide pseudo labels where our simple re-ranker or off-the-shelf re-rankers can be used.
We theoretically show that gradient-based optimization provides a generalized version of the Rocchio algorithm for pseudo relevance feedback, which leads us to develop different variants of \ours.
{Experiments show that our test-time query optimization largely improves the retrieval accuracy on multiple open-domain QA datasets in various settings while being more efficient than traditional re-ranking methods.}

\clearpage
\section*{Limitations}

In this paper, we focus on the end-to-end accuracy and passage retrieval accuracy for open-domain QA.
We have also experimented on the BEIR benchmark~\citep{thakur2021beir} to evaluate our method in the zero-shot document retrieval task.
{Overall, we obtained 48.1\% macro-averaged NDCG@10 compared to 47.8\% by the re-ranking method.
{For some tasks, \ours{} obtains significant improvements with a pre-trained document retriever~\citep{Hofsttter2021EfficientlyTA}.
For example, \ours{} improves the baseline retriever by 11.6\% and 23.8\% NDCG@10 on BioASQ and TREC-COVID, respectively, while also outperforming the re-ranker by 2.1\% and 2.4\% NDCG@10.}
We plan to better understand why \ours{} performs better specifically on these tasks and further improve it.
}

{\ours{} also requires a set of validation examples for hyperparameter selection.}
{While we only used in-domain validation examples for \ours, which were also adopted when training re-rankers, we observed some performance variances depending on the hyperparameters.}
We hope to tackle this issue with better optimization in the future.

\section*{Acknowledgements}

{We thank Zexuan Zhong, Mengzhou Xia, Howard Yen, and the anonymous reviewers for their helpful feedback.
This work was supported in part by the ICT Creative Consilience program (IITP-2023-2020-0-01819) supervised by the IITP (Institute for Information \& communications Technology Planning \& Evaluation), National Research Foundation of Korea (NRF-2023R1A2C3004176), and the Hyundai Motor Chung Mong-Koo Foundation.}

\bibliography{custom}

\begin{thebibliography}{42}
\expandafter\ifx\csname natexlab\endcsname\relax\def\natexlab#1{#1}\fi

\bibitem[{Baudi{\v{s}} and {\v{S}}ediv{\`y}(2015)}]{baudivs2015modeling}
Petr Baudi{\v{s}} and Jan {\v{S}}ediv{\`y}. 2015.
\newblock \href {https://doi.org/10.1007/978-3-319-24027-5_20} {Modeling of the
  question answering task in the yodaqa system}.
\newblock In \emph{International Conference of the cross-language evaluation
  Forum for European languages}, pages 222--228. Springer.

\bibitem[{Baumg{\"a}rtner et~al.(2022)Baumg{\"a}rtner, Ribeiro, Reimers, and
  Gurevych}]{baumgartner-etal-2022-incorporating}
Tim Baumg{\"a}rtner, Leonardo F.~R. Ribeiro, Nils Reimers, and Iryna Gurevych.
  2022.
\newblock \href {https://aclanthology.org/2022.emnlp-main.614} {Incorporating
  relevance feedback for information-seeking retrieval using few-shot document
  re-ranking}.
\newblock In \emph{Proceedings of the 2022 Conference on Empirical Methods in
  Natural Language Processing}, pages 8988--9005.

\bibitem[{Berant et~al.(2013)Berant, Chou, Frostig, and
  Liang}]{berant2013semantic}
Jonathan Berant, Andrew Chou, Roy Frostig, and Percy Liang. 2013.
\newblock \href {https://aclanthology.org/D13-1160} {Semantic parsing on
  {F}reebase from question-answer pairs}.
\newblock In \emph{Proceedings of the 2013 Conference on Empirical Methods in
  Natural Language Processing}, pages 1533--1544.

\bibitem[{Chen et~al.(2017)Chen, Fisch, Weston, and Bordes}]{chen2017reading}
Danqi Chen, Adam Fisch, Jason Weston, and Antoine Bordes. 2017.
\newblock \href {https://doi.org/10.18653/v1/P17-1171} {Reading {W}ikipedia to
  answer open-domain questions}.
\newblock In \emph{Proceedings of the 55th Annual Meeting of the Association
  for Computational Linguistics (Volume 1: Long Papers)}, pages 1870--1879.

\bibitem[{Croft et~al.(2010)Croft, Metzler, and Strohman}]{croft2010search}
W~Bruce Croft, Donald Metzler, and Trevor Strohman. 2010.
\newblock \href {https://doi.org/book/10.5555/1516224} {\emph{Search engines:
  Information retrieval in practice}}, volume 520.
\newblock Addison-Wesley Reading.

\bibitem[{Fajcik et~al.(2021)Fajcik, Docekal, Ondrej, and Smrz}]{fajcik2021r2}
Martin Fajcik, Martin Docekal, Karel Ondrej, and Pavel Smrz. 2021.
\newblock \href {https://doi.org/10.18653/v1/2021.findings-emnlp.73} {R2-d2: A
  modular baseline for open-domain question answering}.
\newblock In \emph{Findings of the Association for Computational Linguistics:
  EMNLP 2021}, pages 854--870.

\bibitem[{Hofst{\"a}tter et~al.(2021)Hofst{\"a}tter, Lin, Yang, Lin, and
  Hanbury}]{Hofsttter2021EfficientlyTA}
Sebastian Hofst{\"a}tter, Sheng-Chieh Lin, Jheng-Hong Yang, Jimmy~J. Lin, and
  Allan Hanbury. 2021.
\newblock \href {https://doi.org/10.48550/arXiv.2104.06967} {Efficiently
  teaching an effective dense retriever with balanced topic aware sampling}.
\newblock \emph{Proceedings of the 44th International ACM SIGIR Conference on
  Research and Development in Information Retrieval}.

\bibitem[{Holtzman et~al.(2020)Holtzman, Buys, Du, Forbes, and
  Choi}]{holtzman2019curious}
Ari Holtzman, Jan Buys, Li~Du, Maxwell Forbes, and Yejin Choi. 2020.
\newblock \href {https://openreview.net/forum?id=rygGQyrFvH} {The curious case
  of neural text degeneration}.
\newblock In \emph{8th International Conference on Learning Representations,
  {ICLR} 2020, Addis Ababa, Ethiopia, April 26-30, 2020}.

\bibitem[{Iyer et~al.(2021)Iyer, Min, Mehdad, and Yih}]{iyer2021reconsider}
Srinivasan Iyer, Sewon Min, Yashar Mehdad, and Wen-tau Yih. 2021.
\newblock \href {https://doi.org/10.18653/v1/2021.naacl-main.100}
  {{RECONSIDER}: Improved re-ranking using span-focused cross-attention for
  open domain question answering}.
\newblock In \emph{Proceedings of the 2021 Conference of the North American
  Chapter of the Association for Computational Linguistics: Human Language
  Technologies}, pages 1280--1287.

\bibitem[{Izacard and Grave(2020)}]{izacard2020distilling}
Gautier Izacard and Edouard Grave. 2020.
\newblock \href {https://doi.org/10.48550/arXiv.2012.04584} {Distilling
  knowledge from reader to retriever for question answering}.
\newblock In \emph{International Conference on Learning Representations}.

\bibitem[{Izacard and Grave(2021)}]{izacard2020leveraging}
Gautier Izacard and Edouard Grave. 2021.
\newblock \href {https://aclanthology.org/2021.eacl-main.74} {Leveraging
  passage retrieval with generative models for open domain question answering}.
\newblock In \emph{Proceedings of the 16th Conference of the European Chapter
  of the Association for Computational Linguistics: Main Volume}, pages
  874--880.

\bibitem[{Joshi et~al.(2017)Joshi, Choi, Weld, and
  Zettlemoyer}]{joshi2017triviaqa}
Mandar Joshi, Eunsol Choi, Daniel Weld, and Luke Zettlemoyer. 2017.
\newblock \href {https://doi.org/10.18653/v1/P17-1147} {{T}rivia{QA}: A large
  scale distantly supervised challenge dataset for reading comprehension}.
\newblock In \emph{Proceedings of the 55th Annual Meeting of the Association
  for Computational Linguistics (Volume 1: Long Papers)}, pages 1601--1611.

\bibitem[{Karpukhin et~al.(2020)Karpukhin, Oguz, Min, Lewis, Wu, Edunov, Chen,
  and Yih}]{karpukhin2020dense}
Vladimir Karpukhin, Barlas Oguz, Sewon Min, Patrick Lewis, Ledell Wu, Sergey
  Edunov, Danqi Chen, and Wen-tau Yih. 2020.
\newblock \href {https://doi.org/10.18653/v1/2020.emnlp-main.550} {Dense
  passage retrieval for open-domain question answering}.
\newblock In \emph{Proceedings of the 2020 Conference on Empirical Methods in
  Natural Language Processing (EMNLP)}, pages 6769--6781.

\bibitem[{Khattab et~al.(2021)Khattab, Potts, and
  Zaharia}]{khattab2021relevance}
Omar Khattab, Christopher Potts, and Matei Zaharia. 2021.
\newblock \href {https://doi.org/10.1162/tacl_a_00405} {Relevance-guided
  supervision for {O}pen{QA} with {C}ol{BERT}}.
\newblock \emph{Transactions of the Association for Computational Linguistics},
  9:929--944.

\bibitem[{Khattab and Zaharia(2020)}]{khattab2020colbert}
Omar Khattab and Matei Zaharia. 2020.
\newblock \href {https://doi.org/10.1145/3397271.3401075} {Colbert: Efficient
  and effective passage search via contextualized late interaction over
  {BERT}}.
\newblock In \emph{Proceedings of the 43rd International {ACM} {SIGIR}
  conference on research and development in Information Retrieval, {SIGIR}
  2020, Virtual Event, China, July 25-30, 2020}, pages 39--48.

\bibitem[{Krause et~al.(2019)Krause, Kahembwe, Murray, and
  Renals}]{krause2019dynamic}
Ben Krause, Emmanuel Kahembwe, Iain Murray, and Steve Renals. 2019.
\newblock \href {https://arxiv.org/abs/1904.08378} {Dynamic evaluation of
  transformer language models}.
\newblock \emph{arXiv preprint arXiv:1904.08378}.

\bibitem[{Kwiatkowski et~al.(2019)Kwiatkowski, Palomaki, Redfield, Collins,
  Parikh, Alberti, Epstein, Polosukhin, Devlin, Lee, Toutanova, Jones, Kelcey,
  Chang, Dai, Uszkoreit, Le, and Petrov}]{kwiatkowski2019natural}
Tom Kwiatkowski, Jennimaria Palomaki, Olivia Redfield, Michael Collins, Ankur
  Parikh, Chris Alberti, Danielle Epstein, Illia Polosukhin, Jacob Devlin,
  Kenton Lee, Kristina Toutanova, Llion Jones, Matthew Kelcey, Ming-Wei Chang,
  Andrew~M. Dai, Jakob Uszkoreit, Quoc Le, and Slav Petrov. 2019.
\newblock \href {https://doi.org/10.1162/tacl_a_00276} {Natural questions: A
  benchmark for question answering research}.
\newblock \emph{Transactions of the Association for Computational Linguistics},
  7:452--466.

\bibitem[{Lavrenko and Croft(2001)}]{lavrenko2001relevance}
Victor Lavrenko and W~Bruce Croft. 2001.
\newblock \href {https://doi.org/10.1145/383952.383972} {Relevance based
  language models}.
\newblock In \emph{Proceedings of the 24th annual international ACM SIGIR
  conference on Research and development in information retrieval}, pages
  120--127.

\bibitem[{Lee et~al.(2021{\natexlab{a}})Lee, Sung, Kang, and
  Chen}]{lee2021learning}
Jinhyuk Lee, Mujeen Sung, Jaewoo Kang, and Danqi Chen. 2021{\natexlab{a}}.
\newblock \href {https://doi.org/10.18653/v1/2021.acl-long.518} {Learning dense
  representations of phrases at scale}.
\newblock In \emph{Proceedings of the 59th Annual Meeting of the Association
  for Computational Linguistics and the 11th International Joint Conference on
  Natural Language Processing (Volume 1: Long Papers)}, pages 6634--6647.

\bibitem[{Lee et~al.(2021{\natexlab{b}})Lee, Wettig, and Chen}]{lee2021phrase}
Jinhyuk Lee, Alexander Wettig, and Danqi Chen. 2021{\natexlab{b}}.
\newblock \href {https://doi.org/10.18653/v1/2021.emnlp-main.297} {Phrase
  retrieval learns passage retrieval, too}.
\newblock In \emph{Proceedings of the 2021 Conference on Empirical Methods in
  Natural Language Processing}, pages 3661--3672.

\bibitem[{Lee et~al.(2019)Lee, Chang, and Toutanova}]{lee2019latent}
Kenton Lee, Ming-Wei Chang, and Kristina Toutanova. 2019.
\newblock \href {https://doi.org/10.18653/v1/P19-1612} {Latent retrieval for
  weakly supervised open domain question answering}.
\newblock In \emph{Proceedings of the 57th Annual Meeting of the Association
  for Computational Linguistics}, pages 6086--6096.

\bibitem[{Lewis et~al.(2022)Lewis, Oguz, Xiong, Petroni, Yih, and
  Riedel}]{lewis-etal-2022-boosted}
Patrick Lewis, Barlas Oguz, Wenhan Xiong, Fabio Petroni, Scott Yih, and
  Sebastian Riedel. 2022.
\newblock \href {https://doi.org/10.18653/v1/2022.naacl-main.226} {Boosted
  dense retriever}.
\newblock In \emph{Proceedings of the 2022 Conference of the North American
  Chapter of the Association for Computational Linguistics: Human Language
  Technologies}, pages 3102--3117.

\bibitem[{Lewis et~al.(2021)Lewis, Stenetorp, and Riedel}]{lewis2020question}
Patrick Lewis, Pontus Stenetorp, and Sebastian Riedel. 2021.
\newblock \href {https://doi.org/10.18653/v1/2021.eacl-main.86} {Question and
  answer test-train overlap in open-domain question answering datasets}.
\newblock In \emph{Proceedings of the 16th Conference of the European Chapter
  of the Association for Computational Linguistics: Main Volume}, pages
  1000--1008.

\bibitem[{Li et~al.(2018)Li, Sun, He, Wang, Hui, Yates, Sun, and
  Xu}]{li2018nprf}
Canjia Li, Yingfei Sun, Ben He, Le~Wang, Kai Hui, Andrew Yates, Le~Sun, and
  Jungang Xu. 2018.
\newblock \href {https://doi.org/10.18653/v1/D18-1478} {{NPRF}: A neural pseudo
  relevance feedback framework for ad-hoc information retrieval}.
\newblock In \emph{Proceedings of the 2018 Conference on Empirical Methods in
  Natural Language Processing}, pages 4482--4491.

\bibitem[{Li et~al.(2021)Li, Mourad, Zhuang, Koopman, and
  Zuccon}]{li2021pseudo}
Hang Li, Ahmed Mourad, Shengyao Zhuang, Bevan Koopman, and Guido Zuccon. 2021.
\newblock \href {https://arxiv.org/abs/2108.11044} {Pseudo relevance feedback
  with deep language models and dense retrievers: Successes and pitfalls}.
\newblock \emph{Journal of ACM Transactions on Information Systems}.

\bibitem[{Liu et~al.(2019)Liu, Ott, Goyal, Du, Joshi, Chen, Levy, Lewis,
  Zettlemoyer, and Stoyanov}]{liu2019roberta}
Yinhan Liu, Myle Ott, Naman Goyal, Jingfei Du, Mandar Joshi, Danqi Chen, Omer
  Levy, Mike Lewis, Luke Zettlemoyer, and Veselin Stoyanov. 2019.
\newblock \href {https://arxiv.org/abs/1907.11692} {Roberta: A robustly
  optimized bert pretraining approach}.
\newblock \emph{arXiv preprint arXiv:1907.11692}.

\bibitem[{Mao et~al.(2021)Mao, He, Liu, Shen, Gao, Han, and
  Chen}]{mao2020generation}
Yuning Mao, Pengcheng He, Xiaodong Liu, Yelong Shen, Jianfeng Gao, Jiawei Han,
  and Weizhu Chen. 2021.
\newblock \href {https://doi.org/10.18653/v1/2021.acl-long.316}
  {Generation-augmented retrieval for open-domain question answering}.
\newblock In \emph{Proceedings of the 59th Annual Meeting of the Association
  for Computational Linguistics and the 11th International Joint Conference on
  Natural Language Processing (Volume 1: Long Papers)}, pages 4089--4100.

\bibitem[{Nogueira and Cho(2019)}]{nogueira2019passage}
Rodrigo Nogueira and Kyunghyun Cho. 2019.
\newblock \href {https://arxiv.org/abs/1901.04085} {Passage re-ranking with
  bert}.
\newblock \emph{arXiv preprint arXiv:1901.04085}.

\bibitem[{Rajpurkar et~al.(2016)Rajpurkar, Zhang, Lopyrev, and
  Liang}]{rajpurkar2016squad}
Pranav Rajpurkar, Jian Zhang, Konstantin Lopyrev, and Percy Liang. 2016.
\newblock \href {https://doi.org/10.18653/v1/D16-1264} {{SQ}u{AD}: 100,000+
  questions for machine comprehension of text}.
\newblock In \emph{Proceedings of the 2016 Conference on Empirical Methods in
  Natural Language Processing}, pages 2383--2392.

\bibitem[{Ram et~al.(2022)Ram, Shachaf, Levy, Berant, and
  Globerson}]{ram-etal-2022-learning}
Ori Ram, Gal Shachaf, Omer Levy, Jonathan Berant, and Amir Globerson. 2022.
\newblock \href {https://doi.org/10.18653/v1/2022.naacl-main.193} {Learning to
  retrieve passages without supervision}.
\newblock In \emph{Proceedings of the 2022 Conference of the North American
  Chapter of the Association for Computational Linguistics: Human Language
  Technologies}, pages 2687--2700.

\bibitem[{Ren et~al.(2021)Ren, Qu, Liu, Zhao, She, Wu, Wang, and
  Wen}]{ren-etal-2021-rocketqav2}
Ruiyang Ren, Yingqi Qu, Jing Liu, Wayne~Xin Zhao, QiaoQiao She, Hua Wu, Haifeng
  Wang, and Ji-Rong Wen. 2021.
\newblock \href {https://doi.org/10.18653/v1/2021.emnlp-main.224}
  {{R}ocket{QA}v2: A joint training method for dense passage retrieval and
  passage re-ranking}.
\newblock In \emph{Proceedings of the 2021 Conference on Empirical Methods in
  Natural Language Processing}, pages 2825--2835.

\bibitem[{Rocchio(1971)}]{rocchio1971relevance}
Joseph Rocchio. 1971.
\newblock Relevance feedback in information retrieval.
\newblock \emph{The Smart retrieval system-experiments in automatic document
  processing}, pages 313--323.

\bibitem[{Rosa et~al.(2022)Rosa, Bonifacio, Jeronymo, Abonizio, Fadaee, Lotufo,
  and Nogueira}]{rosa2022defense}
Guilherme Rosa, Luiz Bonifacio, Vitor Jeronymo, Hugo Abonizio, Marzieh Fadaee,
  Roberto Lotufo, and Rodrigo Nogueira. 2022.
\newblock \href {https://doi.org/10.48550/arXiv.2212.06121} {In defense of
  cross-encoders for zero-shot retrieval}.
\newblock \emph{arXiv preprint arXiv:2212.06121}.

\bibitem[{Sciavolino et~al.(2021)Sciavolino, Zhong, Lee, and
  Chen}]{sciavolino2021simple}
Christopher Sciavolino, Zexuan Zhong, Jinhyuk Lee, and Danqi Chen. 2021.
\newblock \href {https://doi.org/10.18653/v1/2021.emnlp-main.496} {Simple
  entity-centric questions challenge dense retrievers}.
\newblock In \emph{Proceedings of the 2021 Conference on Empirical Methods in
  Natural Language Processing}, pages 6138--6148.

\bibitem[{Seo et~al.(2019)Seo, Lee, Kwiatkowski, Parikh, Farhadi, and
  Hajishirzi}]{seo2019real}
Minjoon Seo, Jinhyuk Lee, Tom Kwiatkowski, Ankur Parikh, Ali Farhadi, and
  Hannaneh Hajishirzi. 2019.
\newblock \href {https://doi.org/10.18653/v1/P19-1436} {Real-time open-domain
  question answering with dense-sparse phrase index}.
\newblock In \emph{Proceedings of the 57th Annual Meeting of the Association
  for Computational Linguistics}, pages 4430--4441.

\bibitem[{Singh et~al.(2021)Singh, Reddy, Hamilton, Dyer, and
  Yogatama}]{singh2021end}
Devendra Singh, Siva Reddy, Will Hamilton, Chris Dyer, and Dani Yogatama. 2021.
\newblock \href {https://doi.org/10.48550/arXiv.2106.05346} {End-to-end
  training of multi-document reader and retriever for open-domain question
  answering}.
\newblock \emph{Advances in Neural Information Processing Systems},
  34:25968--25981.

\bibitem[{Thakur et~al.(2021)Thakur, Reimers, R{\"u}ckl{\'e}, Srivastava, and
  Gurevych}]{thakur2021beir}
Nandan Thakur, Nils Reimers, Andreas R{\"u}ckl{\'e}, Abhishek Srivastava, and
  Iryna Gurevych. 2021.
\newblock \href {https://openreview.net/forum?id=wCu6T5xFjeJ} {{BEIR}: A
  heterogeneous benchmark for zero-shot evaluation of information retrieval
  models}.
\newblock In \emph{Thirty-fifth Conference on Neural Information Processing
  Systems Datasets and Benchmarks Track (Round 2)}.

\bibitem[{Wang et~al.(2021)Wang, Macdonald, Tonellotto, and
  Ounis}]{wang2021pseudo}
Xiao Wang, Craig Macdonald, Nicola Tonellotto, and Iadh Ounis. 2021.
\newblock \href {https://doi.org/10.48550/arXiv.2106.11251} {Pseudo-relevance
  feedback for multiple representation dense retrieval}.
\newblock In \emph{Proceedings of the 2021 ACM SIGIR International Conference
  on Theory of Information Retrieval}, pages 297--306.

\bibitem[{Wolf et~al.(2020)Wolf, Debut, Sanh, Chaumond, Delangue, Moi, Cistac,
  Rault, Louf, Funtowicz, Davison, Shleifer, von Platen, Ma, Jernite, Plu, Xu,
  Le~Scao, Gugger, Drame, Lhoest, and Rush}]{wolf2020transformers}
Thomas Wolf, Lysandre Debut, Victor Sanh, Julien Chaumond, Clement Delangue,
  Anthony Moi, Pierric Cistac, Tim Rault, Remi Louf, Morgan Funtowicz, Joe
  Davison, Sam Shleifer, Patrick von Platen, Clara Ma, Yacine Jernite, Julien
  Plu, Canwen Xu, Teven Le~Scao, Sylvain Gugger, Mariama Drame, Quentin Lhoest,
  and Alexander Rush. 2020.
\newblock \href {https://doi.org/10.18653/v1/2020.emnlp-demos.6} {Transformers:
  State-of-the-art natural language processing}.
\newblock In \emph{Proceedings of the 2020 Conference on Empirical Methods in
  Natural Language Processing: System Demonstrations}, pages 38--45.

\bibitem[{Xiong et~al.(2020)Xiong, Xiong, Li, Tang, Liu, Bennett, Ahmed, and
  Overwijk}]{xiong2020approximate}
Lee Xiong, Chenyan Xiong, Ye~Li, Kwok-Fung Tang, Jialin Liu, Paul~N Bennett,
  Junaid Ahmed, and Arnold Overwijk. 2020.
\newblock \href {https://doi.org/https://doi.org/10.48550/arXiv.2007.00808}
  {Approximate nearest neighbor negative contrastive learning for dense text
  retrieval}.
\newblock In \emph{International Conference on Learning Representations}.

\bibitem[{Yu et~al.(2021)Yu, Xiong, and Callan}]{yu2021improving}
HongChien Yu, Chenyan Xiong, and Jamie Callan. 2021.
\newblock \href {https://doi.org/10.48550/arXiv.2108.13454} {Improving query
  representations for dense retrieval with pseudo relevance feedback}.
\newblock In \emph{Proceedings of the 30th ACM International Conference on
  Information \& Knowledge Management}, pages 3592--3596.

\bibitem[{Zamani et~al.(2018)Zamani, Dehghani, Croft, Learned{-}Miller, and
  Kamps}]{zamani2018neural}
Hamed Zamani, Mostafa Dehghani, W.~Bruce Croft, Erik~G. Learned{-}Miller, and
  Jaap Kamps. 2018.
\newblock \href {https://doi.org/10.1145/3269206.3271800} {From neural
  re-ranking to neural ranking: Learning a sparse representation for inverted
  indexing}.
\newblock In \emph{Proceedings of the 27th {ACM} International Conference on
  Information and Knowledge Management, 2018}, pages 497--506.

\end{thebibliography}
\bibliographystyle{acl_natbib}

\clearpage
\newpage
\appendix

\section{Derivation of the Gradient for \ours$_\text{hard}$}\label{sec:apdx-mml}

\begin{proof}
We compute the gradient of $\mathcal{L}_\text{hard}(\mathbf{q}_{t}, \mathcal{C}_{1:k}^{q_t})$ in Eq.~\eqref{eq:tql-sgd} with respect to the query representation $\mathbf{q}_t$.
Denoting $\sum_{\tilde{c}}P_k(\tilde{c}|q_t)$ as $Z$, the gradient is:\vspace{-0.2cm}
\newcommand{\Z}{$\sum_{\tilde{c}}{P_k(\tilde{c}|q)}$}

\small
\begin{equation*}
\begin{split}
    &\frac{\partial \mathcal{L}_\text{hard}(\mathbf{q}_{t}, \mathcal{C}_{1:k}^{q_t})}{\partial \mathbf{q}_t} = \frac{\partial \mathcal{L}_\text{hard}(\mathbf{q}_{t}, \mathcal{C}_{1:k}^{q_t})}{\partial Z} \frac{\partial Z}{\partial \mathbf{q}_t}\\
    &= -\frac{1}{Z}\sum_{\tilde{c}}\frac{\partial P_k(\tilde{c}|q_t)}{\partial \mathbf{q}_t} \\
    &= -\frac{1}{Z}\sum_{\tilde{c}} \sum_{i=1}^k\frac{\partial P_k(\tilde{c}|q_t)}{\partial \mathbf{q}_t^\top\mathbf{c}_i}\frac{\partial \mathbf{q}_t^\top\mathbf{c}_i}{\partial \mathbf{q}_t} \\
    &= -\frac{1}{Z}\sum_{\tilde{c}} \sum_{i=1}^k (\delta[c_i = \tilde{c}] - P_k(c_i|q_t))P_k(\tilde{c}|q_t) \mathbf{c}_i \\
    &= -\sum_{\tilde{c}} \big[ P(\tilde{c}|q_t) \sum_{i=1}^k (\delta[c_i = \tilde{c}] - P_k(c_i|q_t)) \mathbf{c}_i \big]\\
    &= -\sum_{\tilde{c}} P(\tilde{c}|q_t)\big[(1 - P_k(\tilde{c}|q_t)) \tilde{\mathbf{c}} - \smashoperator{\sum_{c\in \mathcal{C}_{1:k}^{q_t}, c\not=\tilde{c}}} P_k(c|q_t)\mathbf{c} \big]\\
    &= -\sum_{\tilde{c}} P(\tilde{c}|q_t)(1 - P_k(\tilde{c}|q_t)) \tilde{\mathbf{c}} \\
    &\quad+ \sum_{\tilde{c}} \big[ P(\tilde{c}|q_t) \smashoperator{\sum_{c\in \mathcal{C}_{1:k}^{q_t}, c\not=\tilde{c}}} P_k(c|q_t)\mathbf{c} \big]
\end{split}
\end{equation*}\vspace{-0.3cm}

\normalsize
Then, we have:\vspace{-0.3cm}

\begin{align*}
    g(\mathbf{q}_{t},&~\mathcal{C}_{1:k}^{q_t}) = \mathbf{q}_{t} -
    \eta \frac{\partial \mathcal{L}_\text{hard}(\mathbf{q}_{t}, \mathcal{C}_{1:k}^{q_t})}{\partial \mathbf{q}_{t}} \\
    =\mathbf{q}_t &+ \eta \sum_{\tilde{c}} P(\tilde{c}|q_t)(1 - P_k(\tilde{c}|q_t)) \tilde{\mathbf{c}} \\
    &-\eta \sum_{\tilde{c}} \big[ P(\tilde{c}|q_t) \smashoperator{\sum_{c\in \mathcal{C}_{1:k}^{q_t}, c\not=\tilde{c}}} P_k(c|q_t)\mathbf{c} \big].
    \end{align*}
\end{proof}

\vspace{-0.4cm}

\section{Derivation of the Gradient for \ours$_\text{soft}$}\label{sec:apdx-kl}
\begin{proof}
\normalsize
We compute the gradient of $\mathcal{L}_\text{soft}(\mathbf{q}_{t}, \mathcal{C}_{1:k}^{q_t})$ in Eq.~\eqref{eq:tql-kl} with respect to $\mathbf{q}_t$. Denoting $P(c_i=c^*|q_t,\cross)$ as $P_i$, we expand the loss term as:\vspace{-0.3cm}

\small
\begin{align*}
    &\quad\mathcal{L}_\text{soft}(\mathbf{q}_{t},\mathcal{C}_{1:k}^{q_t})=-\smashoperator{\sum_{i=1}^k} P_i \log \frac{P_k(c_i|q_t)}{P_i} \\
    &\quad  =-\sum_{i=1}^k P_i(\mathbf{q}_t^\top \mathbf{c}_i - \log \sum_{j=1}^k \exp(\mathbf{q}_t^\top \mathbf{c}_j)-\log P_i).
\end{align*}

\normalsize
Then, the gradient is:\vspace{-0.2cm}

\small
\begin{align*}
    &\quad\frac{\partial \mathcal{L}_\text{soft}(\mathbf{q}_{t}, \mathcal{C}_{1:k}^{q_t})}{\partial \mathbf{q}_t} \\
    & =-\sum_{i=1}^k P_i \frac{\partial}{\partial \mathbf{q}_t}(\mathbf{q}_t^\top \mathbf{c}_i - \log\sum_{j=1}^k \exp(\mathbf{q}_t^\top \mathbf{c}_j) - \log P_i) \\
\end{align*}
\vspace{-0.3cm}
\begin{align*}
    & =-\sum_{i=1}^k P_i (\mathbf{c}_i - \frac{1}{\sum_{j=1}^k \exp(\mathbf{q}_t^\top \mathbf{c}_j)} \sum_{j=1}^k \mathbf{c}_j \exp(\mathbf{q}_t^\top \mathbf{c}_j)) \\
    & =-\sum_{i=1}^k P_i (\mathbf{c}_i - \sum_{j=1}^k \mathbf{c}_j \frac{\exp(\mathbf{q}_t^\top \mathbf{c}_j)}{\sum_{l=1}^k \exp(\mathbf{q}_t^\top \mathbf{c}_l)}) \\
    & =-\sum_{i=1}^k P_i (\mathbf{c}_i - \sum_{j=1}^k P_k(c_j|q_t)\mathbf{c}_j) \\
    & =-\sum_{i=1}^k P_i \mathbf{c}_i + \sum_{i=1}^k P_k(c_i|q_t)\mathbf{c}_i.
\end{align*}

\normalsize
Putting it all together:\vspace{-0.5cm}

\begin{align*}
    &g(\mathbf{q}_{t}, \mathcal{C}_{1:k}^{q_t}) = \mathbf{q}_{t} -
    \eta \frac{\partial \mathcal{L}_\text{soft}(\mathbf{q}_{t}, \mathcal{C}_{1:k}^{q_t})}{\partial \mathbf{q}_{t}} \\
    &=\mathbf{q}_t + \eta\sum_{i=1}^k P(c_i|q_t, \cross) \mathbf{c}_i - \eta\sum_{i=1}^k P_k(c_i|q_t)\mathbf{c}_i.
\end{align*}

\end{proof}

\section{Relation to the Rocchio Algorithm}\label{sec:apdx-equality}

\begin{proof}
We derive Eq.~\eqref{eq:prf} from Eq.~\eqref{eq:generalized}.
\begin{equation}
\begin{split}
  g(\mathbf{q}_{t}&, \topkret{q_t}) \\
  \quad=\mathbf{q}_t &+ \eta \sum_{\tilde{c}} \pctq(1 - \pkctq) \tilde{\mathbf{c}} \\
  \quad
  &-\eta \sum_{\tilde{c}} \big[ \pctq \smashoperator{\sum_{c \in \mathcal{C}_{1:k}^{q_t}, c\not=\tilde{c}}} \pkcq \mathbf{c} \big] \\
  \quad=\mathbf{q}_t &+ \eta \sum_{i=1}^{k'} \frac{1}{k'}(1 - \frac{1}{k}) \mathbf{c}_i \\
  \quad
  &-\eta \sum_{i=1}^{k'} \big[ \frac{1}{k'} \smashoperator{\sum_{j=1, i\not=j}^{k}} \frac{1}{k} \mathbf{c}_j \big] \\
  \quad=\mathbf{q}_t &+ \eta \frac{k-1}{k'k} \sum_{i=1}^{k'} \mathbf{c}_i \\
  \quad
  &-\eta \frac{1}{k'k} \sum_{i=1}^{k'} \big[  \smashoperator{\sum_{j=1, i\not=j}^{k'}} \mathbf{c}_j + \smashoperator{\sum_{j=k'+1}^{k}} \mathbf{c}_j  \big] \\
    \quad=\mathbf{q}_t &+ \eta \frac{k-1}{k'k} \sum_{i=1}^{k'} \mathbf{c}_i \\
  \quad
  &-\eta \frac{1}{k'k} \big[(k'-1) \sum_{i=1}^{k'} \mathbf{c}_i + k'\smashoperator{\sum_{i=k'+1}^{k}} \mathbf{c}_i  \big] \\
    \quad=\mathbf{q}_t &+ \eta \frac{k-k'}{k'k} \sum_{i=1}^{k'} \mathbf{c}_i -\eta \frac{1}{k} {\sum_{j=k'+1}^{k}} \mathbf{c}_j.  \\
\end{split}
\end{equation}

Then, the equality holds when $\alpha=1$, $\beta=\eta\frac{k-k'}{k}$, and $\gamma=\eta\frac{k-k'}{k}$.
\end{proof}

\section{Implementation Details}\label{sec:apdx-detail}
\paragraph{{Phrase re-ranker}}
{To train a cross-encoder re-ranker for phrase retrieval~(\Cref{sec:labeler}), we first annotate the top 100 retrieved results from DensePhrases.} 
We use three sentences as our context, one that contains a retrieved phrase and the other two that surround it.
This leads to faster inference than using the whole paragraph as input while preserving the performance.
During the 20 epochs of training, we sample positive and negative contexts for every epoch while selecting the best re-reanker based on the validation accuracy of the re-ranker.
We modified the code provided by the Transformers library\footnote{\url{https://github.com/huggingface/transformers/blob/v4.13.0/examples/pytorch/text-classification/run\_glue.py}}~\citep{wolf2020transformers} and used the same hyperparameters as specified in their documentation except for the number of training epochs.
{The ablation study in \Cref{tab:reranker_ablation} shows that we can achieve stronger performance by prepending titles to inputs, using larger language models, using three sentences as our context, and pre-training over reading comprehension datasets.
}
{Using entire paragraphs as input contexts only slightly increases performance compared to using three sentences, but it doubles the query latencies of re-ranking.}

\begin{table}[h]
    \centering
    \resizebox{0.75\columnwidth}{!}{%
    \begin{tabular}{lc}
        \toprule
        &\textsc{NQ} \\
        \midrule
        Phrase re-ranker & 45.4\\
        \quad Without prepending titles & 44.8 \\
        \quad \text{Rl} $\Rightarrow$ \text{Rb} & 43.2\\
        \quad $\text{3} \Rightarrow \text{1 sentence}$ & 43.6\\
        \quad $\text{3} \Rightarrow$ \text{Paragraph$^*$} & \textbf{45.6}\\
        \quad $\text{RC} \Rightarrow$ \text{MNLI pre-training} & 43.8 \\
        \quad $\text{RC} \Rightarrow$ \text{No pre-training} & 42.0\\
        \bottomrule
    \end{tabular}
    }
    \caption{
    {Ablation study of our phrase re-ranker. Rl: RoBERTa-large.
    Rb: RoBERTa-base.
    RC: reading comprehension.
    *: using entire paragraphs as input doubles query latencies.}
    }
    \label{tab:reranker_ablation}
\end{table}

\paragraph{Dense retriever}
We modified the official code of DensePhrases\footnote{\url{https://github.com/princeton-nlp/DensePhrases}}~\citep{lee2021learning} and DPR\footnote{\url{https://github.com/facebookresearch/DPR}}~\citep{karpukhin2020dense} to implement ~\ours~on dense retrievers.
While pre-trained models and indexes of DensePhrases$_\text{multi}$ and DPR$_\text{NQ}$ are publicly available, the indexes of DensePhrases$_\text{NQ}$ and DPR$_\text{multi}$ have not been released as of May 25th, 2022.
When necessary, we reimplemented them to experiment with open-domain QA and passage retrieval in the query distribution shift setting.

\paragraph{Hyperparamter}
When running~\ours~, we use gradient descent with momentum set to 0.99 and use weight decay~$\lambda_\text{decay}=0.01$.
We also perform a linear learning rate scheduling per iteration.
Both the threshold $p$ and temperature $\tau$ for pseudo labels are set to 0.5.
Table~\ref{tab:hyperparameters} lists the hyperparameters that are used differently for each task.
{All hyperparameters of~\ours~were tuned using the in-domain development set.}

\begin{table}[h]
    \centering
    \resizebox{1.0\columnwidth}{!}{%
    \begin{tabular}{lcccc}
    \toprule
    & ODQA & & \multicolumn{2}{c}{Passage Retrieval} \\
    \cmidrule{2-2} \cmidrule{4-5}
    \textbf{Hyperparameter} & DensePhrases & & DensePhrases & DPR \\
    \midrule
    Learning rate $\eta$ & 1.2 & & 1.2 & 0.2 \\
    Max iterations & 3 & & 1 & 1 \\
    Retrieval top-$k$ & 10 & & 100 & 100 \\
    Re-ranker top-$k$ & 10 & & 100 & 100 \\
    Re-ranker $\lambda$ & 0.1 & & 1 & 1 \\
    \bottomrule
    \end{tabular}
    }
    \caption{Hyperparameters of \ours{}for open-domain QA (ODQA) and passage retrieval. 
    }
    \label{tab:hyperparameters}
\end{table}

\section{Data Statistics}\label{sec:apdx-datastats}

\begin{table}[h]
\begin{center}
\centering
\resizebox{0.85\columnwidth}{!}{%
\begin{tabular}{lrrr}
\toprule
\multicolumn{1}{l}{\bf Dataset}  & {\bf Train} & {\bf Dev} & {\bf Test}\\
\midrule
Natural Questions & 79,168 & 8,757 & 3,610 \\
TriviaQA & 78,785 & 8,837 & 11,313 \\
WebQuestions & 3,417 & 361 & 2,032 \\
CuratedTrec & 1,353 & 133 & 694 \\
SQuAD &  78,713 & 8,886 & 10,570 \\
EntityQuestions & - & - & 22,075\\
\bottomrule
\end{tabular}
}
\end{center}
\caption{Statistics of open-domain QA datasets.
}
\label{tab:openqa-data}
\end{table}

\Cref{tab:openqa-data} shows the statistics of the datasets used for end-to-end open-domain QA and passage retrieval tasks. 
{For EntityQuestions, we only use its test set for the query distribution shift evaluation.}

\end{document}